\documentclass[preprints,article,submit,pdftex,moreauthors]{Definitions/mdpi} 
\firstpage{1} 
\pubvolume{1}
\issuenum{1}
\articlenumber{0}
\pubyear{2026}
\copyrightyear{2026}
\datereceived{ } 

\usepackage{algorithm}
\usepackage{algpseudocode}
\usepackage{xcolor}
\usepackage{soul}
\usepackage{framed}
\colorlet{shadecolor}{yellow}

\renewcommand{\linenumbers}[1][]{}

\Title{ConSEAL: Connectivity-Informed Streamline Endpoint Alignment for Diffeomorphic Cortical Registration}

\Author{Yang Xiang $^{1}$, Martin Cole $^{2}$ and Zhengwu Zhang $^{1,}$*}

\AuthorNames{Yang Xiang, Martin Cole and Zhengwu Zhang}

\address{%
$^{1}$ \quad Department of Statistics and Operations Research, The University of North Carolina at Chapel Hill; xya@unc.edu\\
$^{2}$ \quad Department of Psychiatry, University of Rochester; Martin\_Cole@URMC.Rochester.edu}

\corres{Correspondence: zhengwu\_zhang@unc.edu}

\abstract{Cortical surface registration is often driven by local geometric descriptors (e.g., sulcal depth and curvature). While this approach achieves geometric correspondence, it neglects the long-range wiring constraints imposed by white-matter anatomy. Diffusion MRI tractography offers these crucial constraints; however, prior connectivity-informed pipelines typically align precomputed connectivity matrices, making the optimization highly sensitive to connectivity estimation and its resolution. In this paper, we introduce ConSEAL (Connectivity-Informed Streamline Endpoint Alignment), a novel connectivity-based surface registration method that aligns cortical surfaces by operating directly on white-matter fiber-tract endpoints. We model tract endpoints as a point cloud on the product manifold $\Omega \times \Omega$, where $\Omega$ represents the spherical domain of the inflated cortical hemispheres. Our alignment method iteratively (i) computes a small diffeomorphic warp for $\Omega$ by minimizing connectivity mismatch, and (ii) updates the endpoints based on this warp. The method relies on a geometric framework that ensures output warps are diffeomorphisms and has a final goal that optimizes the matching of well-known fiber bundles. Experiments on Human Connectome Project (HCP) data demonstrate that ConSEAL improves tract-level correspondence, achieving higher connectivity-level overlap coefficients on major fiber bundles and stronger robustness across grid resolutions for $\Omega$ compared to state-of-the-art methods such as ENCORE and MSMAll.}


\begin{document}


\section{Introduction}

Establishing accurate point-to-point correspondence between cortical surfaces is a prerequisite for essentially all population-level neuroimaging, and in particular for comparing structural-connectivity (SC) networks across individuals: every group analysis of connectomes presupposes that a given cortical location denotes the same areal and connectional unit in every subject \citep{fischl1999cortical}. Inter-subject alignment is traditionally driven by cortical folding geometry, e.g., sulcal depth and curvature, on a spherical representation of the cortex \citep{fischl1999cortical,yeo2009spherical}, sometimes augmented with cortical thickness \citep{das2009registration} or, as in the widely used MSMAll pipeline, with myelin maps and resting-state functional connectivity \citep{RobinsonMSM,glasser2016multi}. Cortical folding, however, varies markedly across individuals---most severely in the association cortex \citep{mueller2013individual,gratton2018functional}---and is only weakly coupled to the underlying areal and connectional organization, so folding-based alignment frequently fails to bring structurally or functionally homologous regions into register \citep{glasser2016multi}.

\begin{figure}[t]
    \centering
    \includegraphics[width=0.9\columnwidth]{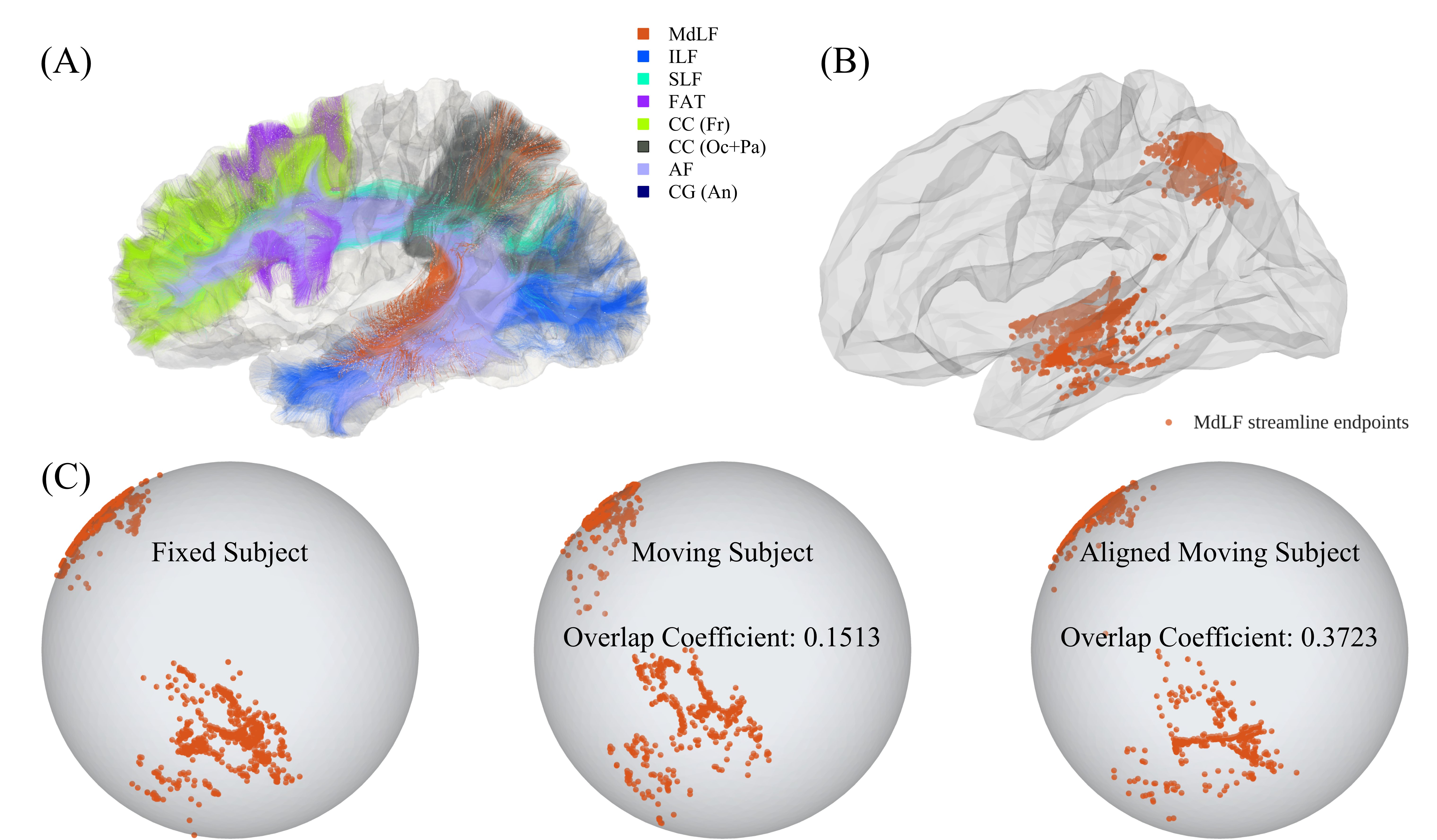}
    \caption{(A) Major streamlines on the cortical white surface. (B) Endpoints of streamlines from the Middle Longitudinal Fasciculus (MdLF) on the cortical white surface. 
     (C): \textbf{Left:} Endpoints from (B) mapped to the spherical domain $\mathbb{S}^2$, serving as the fixed subject (target). \textbf{Middle:} 
    Endpoints from the moving subject (source) before registration. \textbf{Right:}
    Endpoints from the moving subject after alignment to the target using ConSEAL. Note that the direct point-cloud formulation successfully registers the structural constraints without relying on grid-based discretization.}
    \label{fig:endpoint_illustration}
\end{figure}

This misregistration is not merely a geometric inconvenience. When homologous connections are not aligned across subjects, the cross-subject variability of the resulting connectome is inflated by a nuisance component unrelated to genuine individual differences. In principle, such excess edge-level variance can reduce statistical power in group analyses, weaken brain--behavior prediction \citep{smith2015positive}, and---where correspondence is poor---lower the test--retest reliability and individual identifiability of connectomes \citep{finn2015functional,amico2018quest}. Structural connectivity derived from diffusion MRI (dMRI) tractography provides exactly the signal needed to drive such an alignment \citep{cole2025alignment}: tractography reconstructs the long-range white-matter pathways linking distant cortical regions (Figure~\ref{fig:endpoint_illustration}(A)), and because these pathways are strongly constrained by neurodevelopment they remain comparatively stable across individuals \citep{jbabdi2015measuring,osmanliouglu2020connectomic}. Acting as long-range anatomical anchors, connectivity features can disambiguate regions with similar folding but distinct connectivity, improving the alignment of corresponding cortical areas \citep{lazar2010mapping,milano2017extensive,zhang2022quantitative}.

Realizing this promise is methodologically difficult. Many SC-driven methods first convert streamlines into a discretized, precomputed connectivity matrix and then align those matrices \citep{gutman2014registering,zhang2016group,zhou2026deep}. This makes the optimization highly sensitive to connectivity estimation and grid resolution: coarse grids blur fine-scale connectivity, whereas fine grids are extremely sparse, numerically unstable, and computationally costly. Continuous-connectivity formulations remove the explicit resolution dependence \citep{cole2025alignment}, but iteratively warping a continuous, surface-based profile requires repeated re-interpolation and Jacobian-determinant evaluations that introduce numerical diffusion and accumulate error across iterations, degrading the final alignment. Deep-learning registration methods \citep{balakrishnan2019voxelmorph,zhao2021s3reg,cheng2020cortical,li2024josa} are fast at inference but typically rely on explicit regularization or architectural constraints to encourage smooth, invertible deformations rather than guaranteeing them by construction.

To overcome these limitations, we propose ConSEAL (Connectivity-Informed Streamline Endpoint Alignment), a surface-registration framework that operates directly on the exact cortical endpoints of white-matter fiber tracts. We represent each subject's tract endpoints as a point cloud on the product manifold $\Omega \times \Omega$, where $\Omega$ denotes the spherical domain of the inflated cortical hemispheres; this representation jointly encodes endpoint location and connectivity pattern (Figure~\ref{fig:endpoint_illustration}(B,C)). ConSEAL alternates between two steps: (i) estimating a small diffeomorphic deformation of $\Omega$ that minimizes an endpoint-mismatch objective, and (ii) applying that deformation directly to the physical streamline endpoints. Figure~\ref{fig:endpoint_illustration}(C) illustrates the alignment for a single bundle (the MdLF, shown for visualization only---registration always uses all streamline endpoints): the overlap between the target and moving subject's endpoints is markedly improved after alignment.

We summarize our main contributions as follows:
\begin{enumerate}
    \item ConSEAL uses a continuous density proxy solely to compute deformation gradients while applying every update to the discrete endpoint coordinates. The estimated transformation is a strict diffeomorphism by construction, i.e., smooth, invertible, and topology-preserving, without explicit regularization.
    \item Because deformations are applied to endpoints rather than to a grid, ConSEAL bypasses connectivity-matrix construction entirely. This matrix-free design eliminates numerical diffusion and renders the method robust to discretization resolution.
    \item On Human Connectome Project (HCP) data, ConSEAL aligns major white-matter bundles more accurately than state-of-the-art geometric and connectivity-based methods, including MSMAll \citep{RobinsonMSM} and ENCORE \citep{cole2025alignment}.
    \item Going beyond registration accuracy, we show that ConSEAL improves population-level SC network mapping: relative to MSMAll it reduces the cross-subject variance of parcel-wise connectivity while preserving mean connectivity, concentrates its largest deformations in the Limbic and Default Mode networks, and improves prediction of cognitive phenotypes (matrix reasoning and oral reading). These analyses make the contribution to network neuroscience explicit and address the concern that variance reduction alone could be trivial.
\end{enumerate}

The remainder of the paper is organized as follows. Section~\ref{sec:method} presents ConSEAL: the endpoint representation, the Fisher--Rao warping energy, and the iterative alignment, together with the HCP data, preprocessing, and evaluation metrics. Section~\ref{sec:results} reports simulation studies with known ground truth and the HCP analyses, covering bundle-level alignment accuracy and the population-level network results. Section~\ref{sec:discussion} discusses implications and limitations. Full derivations, the complete algorithm, and additional details appear in the Supplementary Material.

\section{Methods}
\label{sec:method}

\begin{figure}[h]
    \centering
    \includegraphics[trim=0 0 0 0, clip, width=0.9\columnwidth]{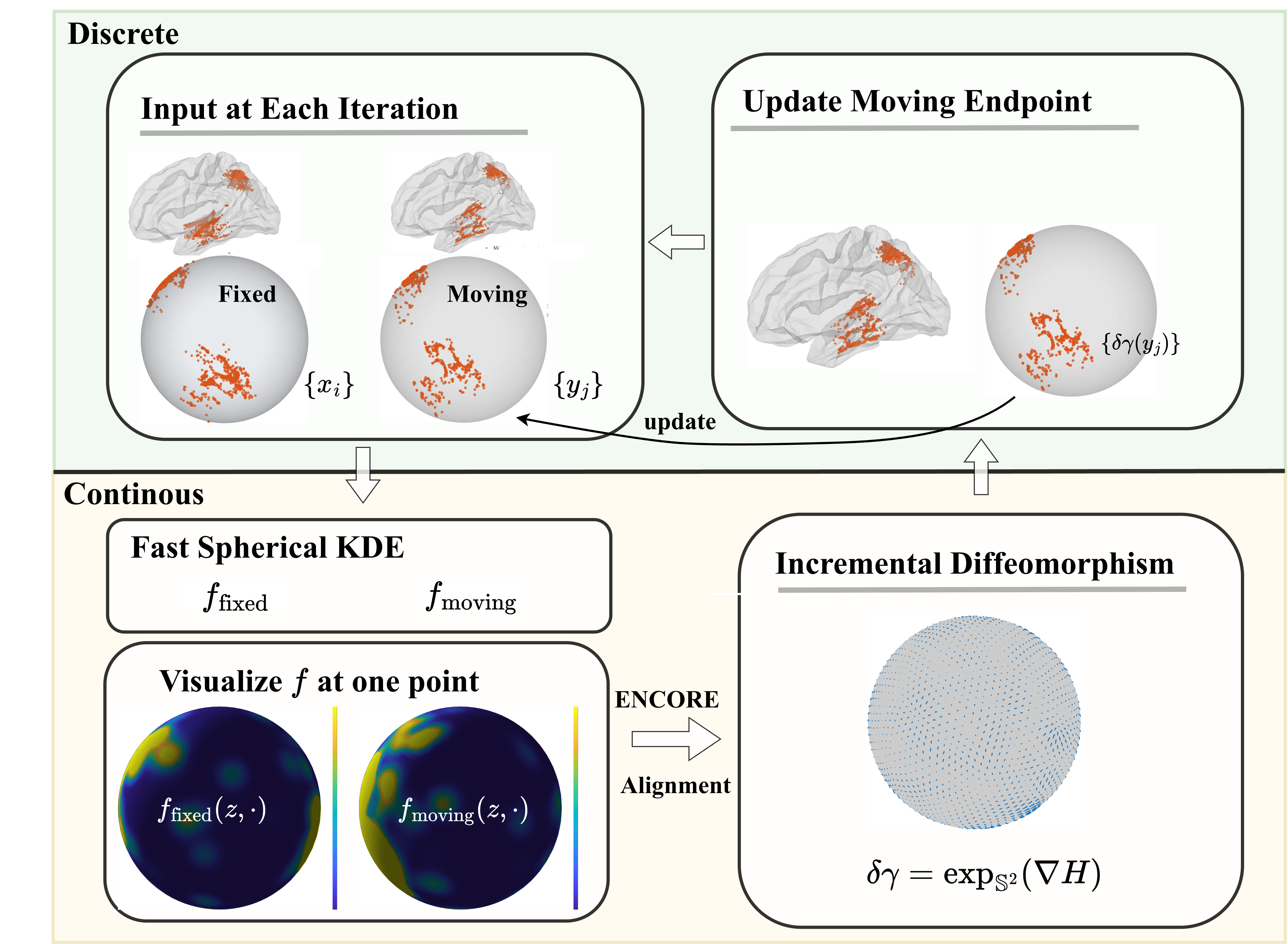}
    \caption{A diagram illustrating the ConSEAL pipeline for white-matter endpoint-based alignment. For visualization, only the left hemisphere and the endpoints of the MdLF fiber bundle are shown. In the full analysis, both hemispheres and all streamline endpoints are used. A fast spherical KDE is used to compute the density proxies. Since the density proxy is defined on a two-dimensional space, we select a random point $z$ and visualize $f(z,\cdot)$ on the sphere. The incremental diffeomorphism $\delta\gamma = \exp_{\mathbb{S}^2}(\nabla H)$, produced by the ENCORE engine \protect\citep{cole2025alignment} using the density proxy, is then applied directly to update the endpoints on the sphere.  Throughout the pipeline, the endpoints of the fixed subject remain unchanged.}
    \label{fig:pipeline}
\end{figure}

We present ConSEAL, a connectivity-driven framework for cortical surface alignment based on streamline endpoints. Given streamline endpoints represented as discrete point clouds on a product manifold, our pipeline bridges the gap between discrete anatomical data and continuous geometric registration through three core iterative steps: (1) projecting these points into a continuous density proxy via a fast spherical Kernel Density Estimation (KDE); (2) applying ENCORE \citep{cole2025alignment}, a diffeomorphic warping framework, to estimate an incremental diffeomorphism that aligns the moving proxy toward the fixed target; and (3) applying the resulting warp directly to the point cloud of the moving subject. This design elegantly separates the domains: step (3) operates on the exact coordinates in the discrete space, whereas steps (1) and (2) compute the gradients in the continuous space. Figure~\ref{fig:pipeline} summarizes the pipeline.

\subsection{Objective and Problem Formulation}
\label{subsec:problem_formulation}

Let $\mathcal{S}_1$ and $\mathcal{S}_2$ be the left and right white-matter cortical surfaces of a subject, and let $\widetilde{\Omega} = \mathcal{S}_1 \cup \mathcal{S}_2$. Both $\mathcal{S}_1$ and $\mathcal{S}_2$ are homeomorphic to the sphere $\mathbb{S}^2$ (Figure~\ref{fig:endpoint_illustration}(B)(C)). Our primary objective is to establish accurate, point-to-point anatomical correspondence for $\widetilde{\Omega}$ across different subjects by aligning the cortical endpoints of white-matter fiber bundles.

To capture white-matter fiber bundle-based structural connectivity, we rely on dMRI tractography, which computationally reconstructs long-range white-matter trajectories known as streamlines \citep{zhang2022quantitative}. Because standard tractography often terminates prematurely in shallow white matter, we utilize the surface-enhanced tractography (SET) algorithm \citep{st2018surface}. Beyond ensuring that reconstructed streamlines terminate exactly on the cortical boundary $\widetilde{\Omega}$, SET mitigates the well-known gyral bias of conventional tractography---the tendency of streamlines to over-terminate on gyral crowns rather than within sulcal banks. This property is especially important for connectivity-driven registration: by reducing termination biases that are themselves tied to cortical folding, SET helps ensure that the endpoint distributions we align reflect genuine structural connectivity rather than artifacts of the very folding pattern we aim to look beyond.

Given this framework, let $(\tilde{p}_1,\tilde{p}_2)$ denote a pair of endpoints belonging to a single streamline, where $\tilde{p}_1,\tilde{p}_2 \in \widetilde{\Omega}$. Through standard parameterization, we map the folded cortical geometry to a spherical domain \citep{fischl1999cortical}. We denote $(p_1,p_2)$ as the mapped image of $(\tilde{p}_1,\tilde{p}_2)$ on the product manifold $\Omega \times \Omega$, where $\Omega = \mathbb{S}_1^2 \cup \mathbb{S}_2^2$ represents the combined left and right spherical spaces. For subject $i$, the set of $N_i$ streamline endpoints mapped onto the sphere is denoted as $\{(p_1^1,p_2^1),\cdots,(p_1^{N_i},p_2^{N_i})\}$.

These extracted point clouds serve as the input for alignment. However, establishing a direct discrete point-matching across subjects is mathematically intractable due to vastly different spatial patterns, varying point densities, and unequal streamline counts ($N_1 \neq N_2$). To overcome this, we elevate the discrete endpoints to a continuous domain, representing them via a smooth spatial density proxy. This continuous proxy acts as an iterative guiding signal: rather than treating it as a ``true" statistical density to be perfectly estimated, ConSEAL leverages it to compute a mathematically rigorous diffeomorphic gradient, which is then applied directly back to align the discrete endpoint point cloud.

\subsection{Data and Preprocessing}
\label{sec:dataset_and_preprocess}
We evaluate ConSEAL using data from the Human Connectome Project (HCP), a large-scale neuroimaging study designed to characterize human brain connectivity in a cohort of healthy adults. The HCP provides multimodal MRI acquisitions, including structural MRI, diffusion MRI (dMRI), and both resting-state and task-based functional MRI, enabling detailed reconstruction of anatomical and functional brain networks.

The dMRI data are processed using the structural connectivity analysis SBCI pipeline \citep{cole2021}. This pipeline reconstructs cortical surfaces and computes their spherical parameterizations via FreeSurfer \citep{fischl1999cortical}. White matter fiber tracts connecting the cortical surfaces are then estimated using SET \citep{st2018surface}, yielding streamline-based representations of structural connectivity. For each subject, approximately $5\times 10^5$ to $10^6$ streamlines are generated. In all subsequent analyses, we retain only the corresponding streamline endpoints, and every subject is registered to a common template subject.

To assess alignment accuracy, we evaluate whether endpoints associated with major white matter fiber bundles exhibit improved cross-subject agreement after registration. Specifically, we quantify the overlap of bundle-specific endpoint distributions across subjects and compare ConSEAL with commonly used registration approaches. The underlying hypothesis is that improved anatomical alignment of connectomes should yield higher inter-subject agreement in endpoint locations, as measured by a connectivity-level overlap coefficient.

\subsection{Continuous Intensity Proxies and Warping Energy}
\label{sec:ConCon_and_energy}

To construct the continuous density proxy, we map the spatial distribution of streamline endpoints into a smooth function through Kernel Density Estimation (KDE) \citep{wang2025riemannian}. Specifically, given the endpoint pairs $\{(p_j^1, p_j^2)_{j=1}^N\} \subseteq \Omega \times \Omega$, we define the density proxy function $f: \Omega \times \Omega \mapsto [0,\infty]$ via a product kernel defined on $\Omega \times \Omega$:
\begin{align}
f(x,y) = \frac{1}{N} \sum_{j=1}^N K_\sigma(x, p_j^1) K_\sigma(y, p_j^2),
\label{eq:kde}
\end{align}
where $K_\sigma(\cdot, \cdot)$ is the fast spherical heat kernel \citep{fischer1984heat,camporesi1990harmonic,nowak2019sharp} evaluated on the spherical domain $\Omega$, with $\sigma > 0$ being the bandwidth parameter. The exact formula of $K_{\sigma}$ is given in Supplementary Section~S2. The function defined in \eqref{eq:kde} is continuous and integrates to one over the product domain, i.e., $\int_{\Omega}\int_{\Omega} f(x,y) dxdy = 1$, so it defines a valid density on $\Omega\times\Omega$. To enforce symmetry, we define the symmetrized density
$
f_{\text{sym}}(x,y) = \frac{1}{2}\big(f(x,y) + f(y,x)\big),
$
and use $f_{\text{sym}}$ as the continuous density proxy in practice. By construction, this guarantees that $f_{\text{sym}}(x,y) = f_{\text{sym}}(y,x)$.

Several techniques are used to make this KDE extremely fast. First, truncating the exponentially decaying high-frequency spectral terms in the spectral expansion of $K_\sigma$ (Supplementary Eq.~(S4)) reduces computational cost. Additionally, setting spatial contributions below a prescribed angular threshold to zero makes the resulting kernel matrix highly sparse.
These optimizations make the KDE computationally efficient \citep{seguin2022connectome}. Moreover,
because the spherical heat kernel admits a spectral representation (Supplementary Eq.~(S4)), its basis functions can be precomputed and reused across different bandwidths $\sigma$, reducing the kernel evaluation to a highly efficient matrix-vector contraction. 
Here, $\sigma$ acts as a resolution parameter: smaller values preserve fine-scale endpoint structures, whereas larger values yield smoother representations that emphasize dominant fiber bundles (see Supplementary Section~S2 for details). The entire KDE computation can be reduced to efficient matrix--vector multiplication, avoiding the need for nested loops.

With each subject represented by a continuous function $f$, alignment reduces to finding a diffeomorphic reparameterization that brings a moving proxy $f_2$ into correspondence with a target proxy $f_1$. While the squared $\mathbb{L}^2$ distance is a natural discrepancy measure, it is famously not invariant under diffeomorphic warping (i.e., $\|f_1 - f_2\| \neq \|f_1 \circ \gamma - f_2 \circ \gamma\|$) \citep{cole2025alignment}. This parameterization dependence can lead to spurious optimization minima driven by coordinate artifacts rather than anatomical differences. 

To restore invariance, we apply the square-root (or $Q$-) mapping \citep{srivastava2010shape}:
\begin{align}
Q(f) = \sqrt{f},
\end{align}
and define a modified group action of $\Gamma_{\Omega}$ (the set of all warping functions on $\Omega$). For $q = Q(f)$ and $\gamma \in \Gamma_{\Omega}$,
\begin{align}
(q * \gamma)(x,y) = q(\gamma(x), \gamma(y))
\sqrt{\det(D\gamma_x)}\sqrt{\det(D\gamma_y)},
\label{eqn:groupaction}
\end{align}
where, $q*\gamma$ represents the warped $q$ function with a warping function $\gamma$, and $\det(D\gamma)$ denotes the Jacobian determinant. These Jacobian factors ensure that normalization is maintained under warping. Under this representation, the $\mathbb{L}^2$ distance becomes invariant \citep{cole2025alignment}, i.e.,  $\|q_1 - q_2\| = \|q_1 * \gamma - q_2 * \gamma\|$. Given a fixed target $q_1$ and a moving $q_2$, we formulate the alignment energy as:
\begin{equation}
\label{eqn:warpenergy}
\begin{aligned}
\inf_{\gamma \in \Gamma_{\Omega}} H(f_1, f_2 * \gamma),
&\quad \text{where} \\
H(f_1, f_2 * \gamma)
&= \iint_{\Omega^2} \Bigl(q_1(x,y)-q_2(\gamma(x),\gamma(y)) \\
&\qquad \cdot \sqrt{\det(D\gamma_x)\det(D\gamma_y)}\Bigr)^2\,dx\,dy.
\end{aligned}
\end{equation}

Note that this formulation inherently provides geometric regularization: extreme local deformations are naturally discouraged by the Jacobian determinants without requiring explicit regularization terms. For the domain $\Omega = \mathbb{S}_1^2 \cup \mathbb{S}_2^2$, refer to Supplementary Eq.~(S2) on how to define $\gamma$ separately on each component.

Moreover, the formulation is inherently inverse-consistent, meaning that $H(f_1,f_2*\gamma) = H(f_1*\gamma^{-1},f_2)$. Because $\Gamma_{\Omega}$ is a group, every mapping $\gamma$ has an inverse $\gamma^{-1}$ that also belongs to $\Gamma_{\Omega}$. Therefore, taking the infimum over all $\gamma \in \Gamma_{\Omega}$ is equivalent to taking the infimum over all inverse mappings. As a result, the alignment energy in \eqref{eqn:warpenergy} is symmetric.

\subsection{Iterative Alignment via Direct Endpoint Updates}
\label{sec:iterative_endpoint}

While the objective in \eqref{eqn:warpenergy} provides a sound theoretical basis for alignment, standard implementations, e.g., ENCORE in \cite{cole2025alignment}, optimize this function by repeatedly warping a discretized connectivity matrix. This grid-based approach introduces severe computational bottlenecks. Warping a grid representation requires evaluating Jacobian determinants at every vertex (i.e., the computation of warped $q$-function in \eqref{eqn:groupaction}), typically via finite differences in tangent spaces. On coarse grids, these estimates are noisy and biased. As the warp is accumulated iteratively, these numerical errors compound, introducing artificial diffusion that steers the optimization away from the true minimizer. Refining the grid reduces bias but incurs a quadratic increase in computational cost due to the product domain $\Omega \times \Omega$.

Our core methodological contribution circumvents these limitations by eliminating the need to repeatedly warp a grid-based representation. By decoupling the gradient calculation from the spatial state update, we alternate between estimating a diffeomorphic warp increment from the continuous proxy and \textit{directly updating the exact physical streamline endpoints}. The algorithm operates as follows:

\begin{enumerate}
    \item \textbf{Continuous Gradient Estimation:} At iteration $k$, we freshly estimate the connectivity density $f_2^{(k)}$ directly from the current spatial locations of the warped point cloud using fast KDE introduced in Section \ref{sec:ConCon_and_energy}. We then compute the gradient of the energy functional $\nabla H$ based on a truncated basis expansion (detailed in Supplementary Section~S1 and \cite{cole2025alignment}).
    \item \textbf{Discrete Endpoint Update:} We generate a small incremental diffeomorphic warp $\delta\gamma$ via the exponential map of the gradient. Crucially, instead of applying this warp to the $Q$ function of density $f_2^{(k)}$ with a Jacobian correction (i.e., Equation \eqref{eqn:groupaction}), \textit{we apply $\delta\gamma$ exclusively to the discrete coordinates of the moving streamline endpoints}. 
\end{enumerate}

By utilizing the continuous density merely as a gradient-generating mechanism, we apply all spatial updates directly to the discrete endpoints, entirely avoiding grid-based interpolation errors. Furthermore, because spatial smoothing and diffeomorphic warping are non-commutative operations ($\mathrm{Smooth}(f * \gamma) \neq \mathrm{Smooth}(f) * \gamma$), iteratively warping a pre-smoothed density grid introduces progressive numerical blurring that degrades the anatomical signal. Warping the exact physical endpoints and re-estimating the density from scratch at each step prevents this systematic drift. By preserving the precise spatial localization of the data throughout the optimization, this direct point-cloud alignment is able to achieve a significantly more accurate correspondence of the structural connectivity patterns generated by white-matter streamlines.

The complete numerical procedure is summarized as Algorithm~S1 in the Supplement. At each iteration, we compute a smooth tangent vector field on the base manifold $\mathbb{S}^2$. The incremental spatial warp is then obtained by applying the Riemannian exponential map of the sphere point-wise to evaluate the displacement along this vector field. In the context of optimization on the space of diffeomorphisms, this point-wise application acts as a retraction. While the exact theoretical flow of a smooth vector field is unconditionally diffeomorphic, this point-wise retraction guarantees a diffeomorphism only for sufficiently small step sizes, which ensures that adjacent geodesic trajectories do not cross. By enforcing a small step size in our gradient descent scheme and terminating after a finite number of iterations, the composition of these incremental updates ensures that the overall output transformation remains diffeomorphic.

\subsection{Evaluation Metrics and Implementation Details}
\label{sec:evaluation}

\subsubsection{Connectivity-Level Overlap Coefficient}
\label{sec:alignment_evaluation}

Evaluating the accuracy of cortical surface alignment is inherently challenging due to the absence of a ground-truth diffeomorphic mapping between subjects. Furthermore, while whole-brain white matter tractography broadly captures the brain's structural wiring, it inevitably produces a mixture of true, well-documented anatomical fiber bundles and erroneous or spurious streamlines \citep{maier2016tractography}. To increase the accuracy and biological validity of our alignment measurement, we evaluate performance by relying exclusively on well-characterized, major fiber tracts. If two subjects are accurately aligned at the SC connectivity level, their corresponding major bundles should occupy highly overlapping regions on the cortical surface. 

To isolate these structures, we employ RecoBundlesX, a multi-atlas white matter bundle segmentation framework, to extract major fiber bundles from individual subjects. RecoBundlesX is an enhanced version of the RecoBundles algorithm \citep{garyfallidis2018recognition}, incorporating multi-parameter fusion to improve bundle segmentation accuracy and robustness. This approach allows for precise and reliable extraction of white matter pathways across subjects. The segmentation procedures were implemented using the Scilpy \citep{st2023bundleseg} pipeline.

To quantify this anatomical agreement, we define the \textbf{Connectivity-level Overlap Coefficient}, which directly measures the overlap of suprathreshold connectivity patterns for these specific bundles after registration. Unlike a pointwise distance metric, the overlap coefficient compares whether the same major connections are structurally present in both subjects after alignment, making it well suited for evaluating whether fiber-tract endpoints have been brought into correspondence. Specifically, we partition the sphere into a discrete triangular mesh $\Delta = \{\Delta_1, \dots, \Delta_K\}$ (using an icosphere grid with $G=4$, $V=2562$, $K = 5120$; see Section~\ref{discretization} for details). Let $C_{a,b}^i$ denote the number of streamlines in subject $i \in \{1,2\}$ whose endpoints fall into the triangle pair $\Delta_a$ and $\Delta_b$. With $n_i$ denoting the total number of streamlines belonging to the specific tract of interest in the subject $i$, we define the overlap coefficient at the threshold $\tau \ge 0$ as:
\begin{align}
\label{eqn:connectivity_dice}
    \mathrm{Overlap} \text{ } \mathrm{Coefficient}(\tau) = \frac{\sum_{a\leq b}\mathbf{1}\left(\frac{C_{a,b}^1}{n_1}>\tau\right) \mathbf{1}\left(\frac{C_{a,b}^2}{n_2}>\tau\right)}{\min\left(\sum_{a\leq b}\mathbf{1}\left(\frac{C_{a,b}^1}{n_1}>\tau\right),\sum_{a\leq b}\mathbf{1}\left(\frac{C_{a,b}^2}{n_2}>\tau\right)\right)}.
\end{align}
This score ranges from 0 to 1. A higher value indicates that the two subjects share more of the same major connections above the threshold. In this way, the connectivity-level overlap coefficient rigorously evaluates whether registration preserves and aligns the endpoints of anatomically meaningful fiber tracts.

Because the estimation of the continuous gradients relies on the diffusion parameter $\sigma$ in \eqref{eq:kde}, the choice of $\sigma$ directly influences the final diffeomorphism produced by ConSEAL. It is therefore important to select an appropriate value. Since our objective is to maximize anatomical correspondence, we choose $\sigma$ based on the connectivity-level overlap coefficient, selecting the value that yields the highest overlap for major fiber tract alignment. We detail the empirical selection of $\sigma$ for real data in Section~\ref{sec:hyper_select}.

\subsubsection{Discretization}
\label{discretization}
Although the continuous intensity proxy function is theoretically defined on every point in $\Omega \times \Omega$, in practice it is evaluated on a discrete set of vertex pairs $\{(x_i, x_j)\}$, where each $(x_i,x_j) \in \Omega\times \Omega$ and the vertices lie on a spherical grid. We adopt an icosphere discretization, obtained by recursively subdividing the faces of an icosahedron and projecting the newly created vertices onto the sphere. The icosphere yields near-uniform triangle areas, which aligns with the discretization assumption of approximately uniform sampling. Moreover, the subdivision process is hierarchical: vertices at lower resolutions form a subset of those at higher resolutions. This property makes it straightforward to switch between resolutions. The number of vertices $V$ in an icosphere grid follows $V = 10*4^G+2$, where $G$ denotes the number of subdivision levels. In this paper, unless otherwise specified, we set $G = 4$, which corresponds to $V = 2562$ vertices. Using a finer grid (e.g., an icosphere with $G = 5$) improves the accuracy of the gradient estimation step, but significantly increases the computational cost.

\subsubsection{Hyperparameter Selection}
\label{sec:hyper_select}
Recall that the bandwidth parameter $\sigma$ in \eqref{eq:kde} controls the smoothness of the KDE estimated from the endpoints. Smaller values of $\sigma$ preserve the discrete structure of the endpoints, yielding sharp, localized peaks, whereas larger values produce smoother estimates. Since our goal is to align endpoints corresponding to major fiber tracts, we select $\sigma$ based on alignment performance on the fiber bundles of interest.

\begin{figure}[t!]
    \centering
\includegraphics[width=0.8\columnwidth]{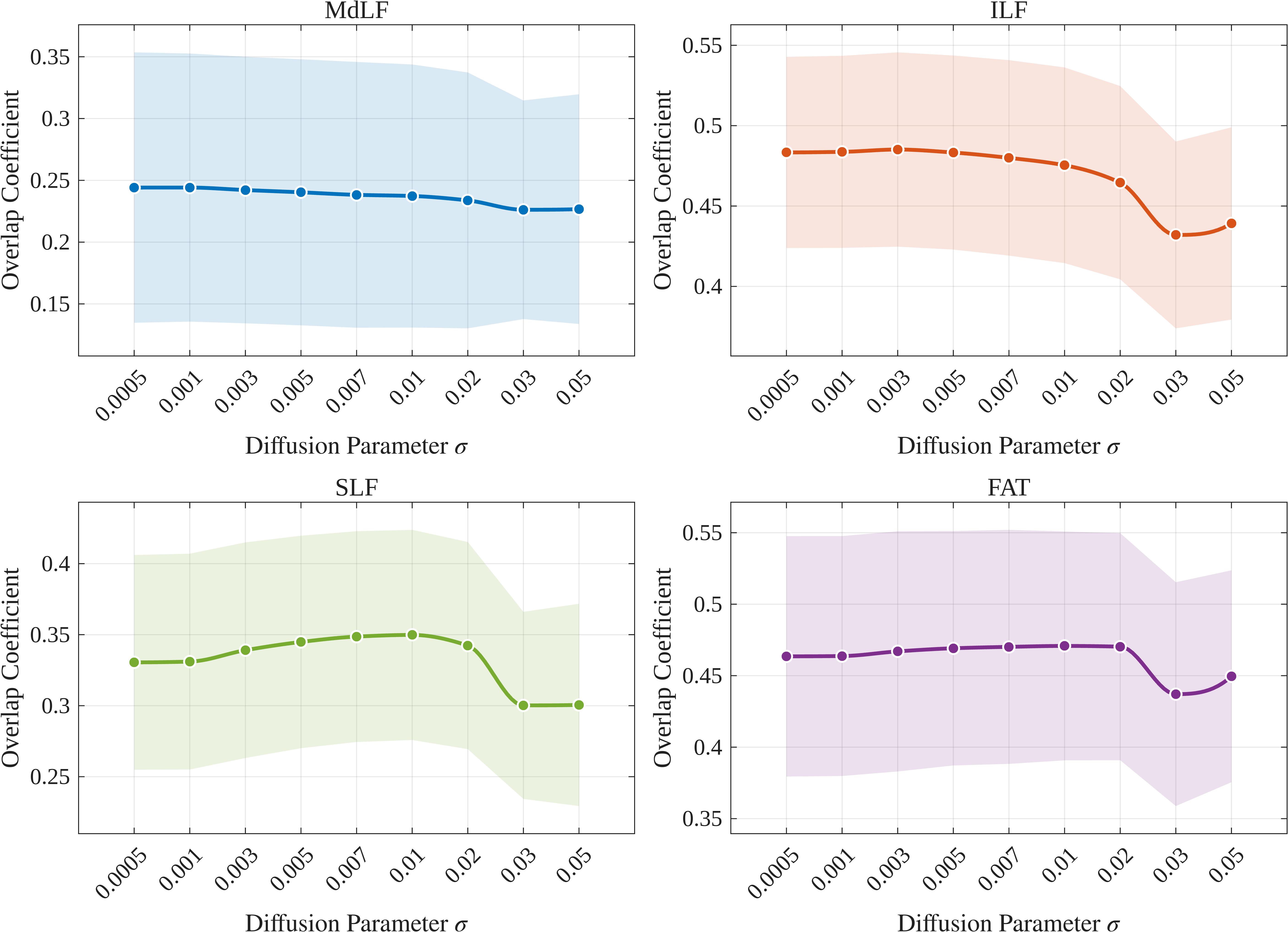}
\vspace{.3cm} \\
\includegraphics[width=0.8\columnwidth]{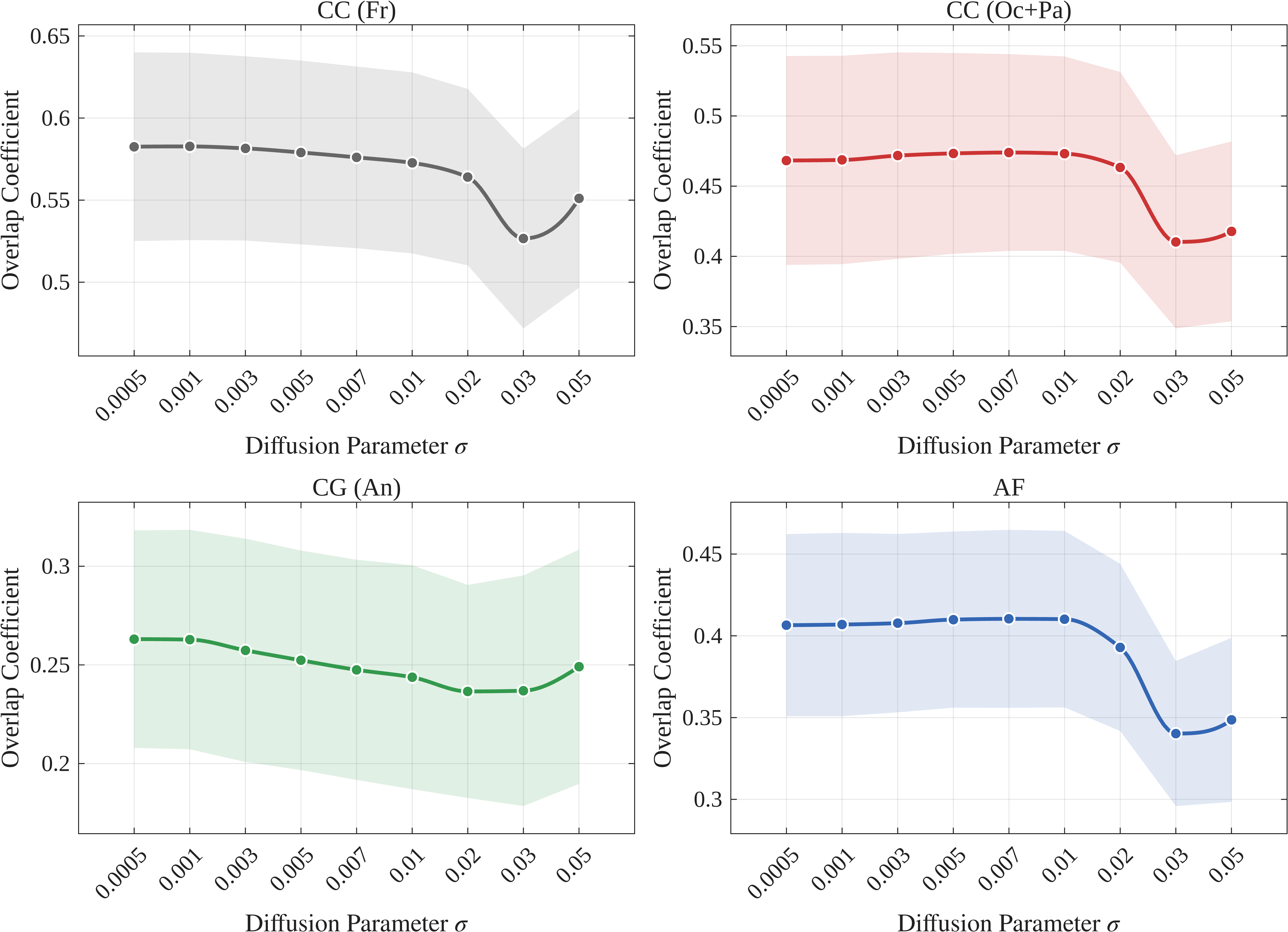}
\caption{Optimization of the spherical heat kernel bandwidth. The connectivity-level overlap coefficient was computed between the template and 50 aligned moving subjects using varying diffusion parameters ($\sigma$). The overall performance demonstrates that $\sigma=0.005$ optimally resolves the spatial density required for successful alignment with ConSEAL.}
    \label{fig:kernel_choice}
\end{figure}

As discussed above, we select the bandwidth parameter $\sigma$ in \eqref{eq:kde} using the connectivity-level overlap coefficient defined in \eqref{eqn:connectivity_dice}. Specifically, we split the dataset into 50 training subjects and use the remaining subjects for testing. We then run ConSEAL on the training subjects over a range of diffusion times $\sigma$, obtaining warped results for each value. For each case, we compute the connectivity-level overlap coefficient between the template and the warped training subjects' streamlines, evaluated across multiple regions of interest. The results are summarized in Figure~\ref{fig:kernel_choice}. Overall, $\sigma = 0.005$ yields the highest weighted overlap coefficient, whether using equal weights across major fiber tracts or weights proportional to the mean number of endpoints in the tracts of interest, compared to other choices. In the subsequent analysis, we use $\sigma = 0.005$, and all analyses are conducted on the test subjects.

\section{Results}
\label{sec:results}

\subsection{Simulation Study with Ground Truth}
\label{sec:simulations}

To evaluate ConSEAL in a controlled setting, we simulate point-cloud data from a known underlying connectivity density function and apply synthetic ground-truth warpings. We demonstrate that ConSEAL accurately recovers the underlying diffeomorphic transformation, even in challenging scenarios where the true deformation is large and the grid resolution is relatively coarse.

We begin by specifying the ground-truth density function used throughout the accuracy and robustness simulations. Let
\begin{align}
h_\kappa(t) := e^{\kappa t}, \quad t \in [-1,1].
\end{align}
Using a von Mises--Fisher (vMF) kernel, we define the joint probability density $f(x,y)$ of the streamline endpoints on the product manifold $\Omega\times\Omega$ as:
\begin{align}
f(x,y) =
\begin{cases}
\displaystyle \alpha \cdot \frac{h_\kappa(x \cdot y)}{2(4\pi)^2  (\sinh \kappa / \kappa)}
& \text{if } x,y \in \mathbb{S}_1^2 \text{ or } x,y \in \mathbb{S}_2^2, 
\\
\displaystyle (1-\alpha) \cdot \frac{1}{2(4\pi)^2}
& \text{if } (x,y) \in \mathbb{S}_1^2 \times \mathbb{S}_2^2 \text{ or } \mathbb{S}_2^2 \times \mathbb{S}_1^2.
\end{cases}
\label{eqn:ground_truth}
\end{align}

This formulation is designed to capture key topological features of true neural wiring. Empirical brain connectivity data typically exhibit stronger and denser within-hemisphere connections than cross-hemisphere connections. To reflect this property, we set the mixture weight $\alpha = 0.85$, thereby assigning significantly greater probability mass to intra-hemispheric interactions. 

Crucially, the parameter $\kappa > 0$ controls the concentration of the vMF kernel, which governs the spatial dispersion of the simulated fiber bundles. In a biological context, $\kappa$ dictates how tightly clustered the simulated endpoints are: a higher $\kappa$ produces sharply localized density peaks (mimicking dense, highly structured major white-matter tracts), whereas a lower $\kappa$ generates more diffuse, unstructured connection patterns. Throughout the simulations, we fix $\kappa = 10$, which provides a realistic balance representing moderately clustered structural connectivity.

\subsubsection{Alignment Accuracy under Large Deformation}

\begin{figure}[h]
    \centering
    \includegraphics[width=0.9\columnwidth]{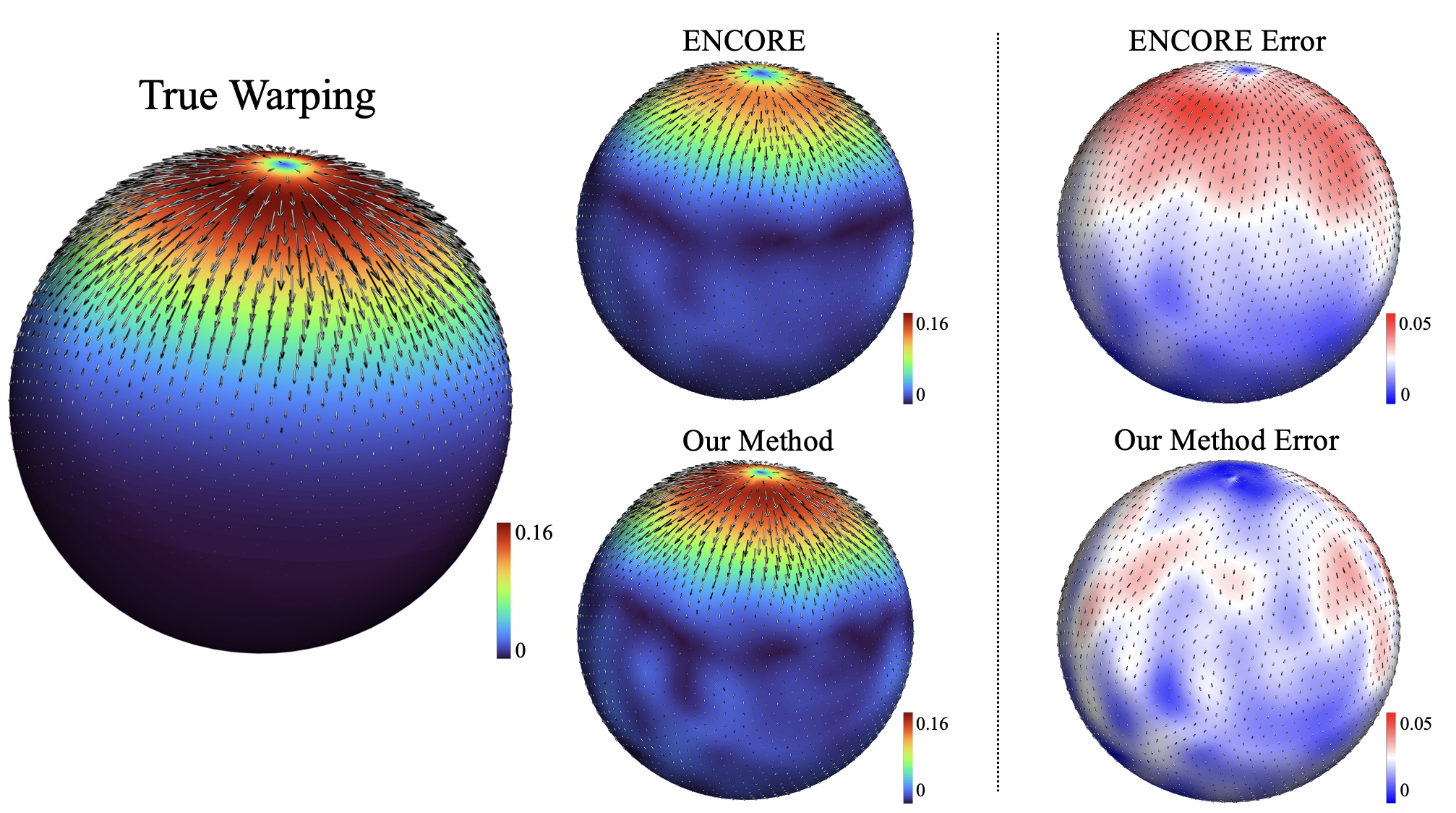}
    \caption{
    \textbf{Left:} The left panel displays the ground-truth warping together with the warpings estimated by the ENCORE algorithm and ConSEAL. The background color represents the magnitude of each warping, as indicated by the colorbar.
    \textbf{Right:} The error of the ENCORE estimate relative to the ground truth is defined as $\log_{\gamma_{\text{ENCORE}}}(\gamma_{\text{truth}})$, the tangent vector at the ENCORE-estimated warping that points toward the ground-truth warping on the sphere. Similarly, the error of the ConSEAL estimate is defined as $\log_{\gamma_{\text{ConSEAL}}}(\gamma_{\text{truth}})$, the tangent vector at the ConSEAL-estimated warping that points toward the ground-truth warping. The background color represents the magnitudes of these displacement vectors, as indicated by the colorbar.
    }
    \label{fig:comparison_encore_endpoint}
\end{figure}

To evaluate the accuracy of ConSEAL under large deformations, we consider a simple spherical example. In this setting, we demonstrate that ConSEAL improves estimation accuracy compared to ENCORE \citep{cole2025alignment} when the ground-truth diffeomorphism induces substantial distortion. Specifically, we use the ground-truth density function $f$ defined in \eqref{eqn:ground_truth} and sample a fixed number of streamlines from this distribution. Given a known ground-truth diffeomorphism $\gamma_{\text{truth}}$ that produces large deformation (see the left panel of Figure~\ref{fig:comparison_encore_endpoint}), we independently draw the same number of streamlines from $f$ and transform their endpoints using $\gamma_{\text{truth}}$. Note that increasing the number of sampled streamlines improves estimation accuracy, as the empirical distribution more closely approximates the true density $f$. To reflect the real-data setting described in Section~\ref{sec:dataset_and_preprocess}, we use $1\times 10^6$ streamlines in this experiment.

We then apply both ENCORE and ConSEAL to estimate the deformation. For a fair comparison, both methods use the same stopping criterion $\epsilon$. For the kernel diffusion bandwidth parameter $\sigma$, we set $\sigma=0.005$ for ConSEAL (see Section~\ref{sec:hyper_select}). For ENCORE, we evaluate multiple kernel scales and report the best result based on mean squared error relative to the ground truth. A visualization is provided in Figure~\ref{fig:comparison_encore_endpoint}. 

The middle column of Figure~\ref{fig:comparison_encore_endpoint} shows the estimated warpings using ENCORE and ConSEAL. The right panel shows the tangent vector fields $\log_{\gamma_{\text {ENCORE}}}\left(\gamma_{\text{truth}}\right)$ and $\log_{\gamma_{\text{ConSEAL}}}\left(\gamma_{\text {truth}}\right)$, which represent the displacements from the ENCORE estimate $\gamma_{\text{ENCORE}}$ to the true warping, and from the ConSEAL estimate $\gamma_{\text{ConSEAL}}$ to the true warping.

ConSEAL achieves higher accuracy than ENCORE. Specifically, ENCORE attains a mean squared error of 0.0359, whereas ConSEAL achieves a lower error of 0.0258. For small angles on the unit sphere, mean squared error is approximately equivalent to mean angular distance, making this comparison geometrically meaningful. Under large deformations, ENCORE struggles to accurately estimate the Jacobian determinant, which limits its ability to recover the true transformation.


\subsubsection{Robustness to Grid Resolution}
\label{robust_to_resolution}
As discussed earlier, the ENCORE algorithm relies on estimating the determinant of the Jacobian matrix, whereas ConSEAL avoids this step. Consequently, changes in the underlying grid used for connectivity estimation have a greater impact on ENCORE than on ConSEAL.

To assess robustness to grid resolution, we apply both ENCORE and ConSEAL on icosphere discretizations with subdivision levels $G = 4$ ($V = 2562$) and $G = 5$ ($V = 10242$). 
The ground-truth connectivity function $f$ is defined in \eqref{eqn:ground_truth}. The corresponding warping function $\gamma$ is shown in Supplementary Figure~S2. Although the ground-truth warping is defined over the entire sphere, the figure displays its icosphere-4 discretization for illustration. The selected ground-truth warping exhibits heterogeneous deformation across the domain, with variations in both magnitude and direction, thereby rendering the estimation of the underlying diffeomorphism a nontrivial task. We evaluate alignment accuracy by measuring the discrepancy between the estimated and true warping using mean angular distance and mean $L_2$ distance. In each experiment, we first independently sample $1\times 10^6$ streamlines for both the fixed and moving subjects according to the ground-truth connectivity $f$. For the fixed subject, the sampled streamline endpoints are then transformed using the prescribed ground-truth warping $\gamma$. Alignment is then performed to register the moving subject to the fixed subject using both \textbf{ENCORE} and \textbf{ConSEAL}. All experiments are conducted on icosphere meshes with 2,562 vertices (level 4) and 10,242 vertices (level 5), using the same stopping criterion for both methods. For ConSEAL, we set the kernel diffusion parameter to $\sigma=0.005$, as discussed in Section~\ref{sec:hyper_select}. For ENCORE, we evaluate multiple kernel scales and report the best-performing result based on the mean angular distance to the ground truth.

To focus on regions with substantial deformation, we restrict evaluation to vertices whose ground-truth warping magnitude is at least the median across all vertices (i.e., the top 50\% by magnitude). On these vertices, we report: (1) the mean angular distance between the true and estimated warping vectors, and (2) the mean $L_2$ distance between the true warped vertices and the estimated warped vertex positions. The results are shown in Table~\ref{tab:warping-performance-compact}. Compared with ENCORE, ConSEAL is more robust to the change from icosphere 4 to icosphere 5, and it produces estimates that are closer to the ground truth.

\begin{table}[ht]
\centering
\caption{Warping estimation performance (evaluation restricted to vertices whose ground-truth warping magnitude $\ge$ median).}
\label{tab:warping-performance-compact}

\small
\setlength{\tabcolsep}{4pt}

\begin{tabularx}{\columnwidth}{@{}l >{\centering\arraybackslash}X c c@{}}
\toprule
Method & Icosphere (vertices) & Mean angular ($^\circ$) & Mean $L_2$ \\
\midrule
ENCORE    & Icosphere 4 (2,562)  & $15.95^\circ$ & 0.0544 \\
ConSEAL   & Icosphere 4 (2,562)  & $10.06^\circ$ & 0.0367 \\
\addlinespace[4pt]
ENCORE    & Icosphere 5 (10,242) & $13.01^\circ$ & 0.0483 \\
ConSEAL   & Icosphere 5 (10,242) & $9.93^\circ$  & 0.0362 \\
\bottomrule
\end{tabularx}

\vspace{4pt}
\footnotesize
\end{table}

\subsection{Real-Data Alignment of Major Fiber Bundles}
\label{sec:real_data}
\label{sec:real_alignment_evaluation}

\begin{figure}[ht]
    \centering
    \includegraphics[width=0.8\columnwidth]{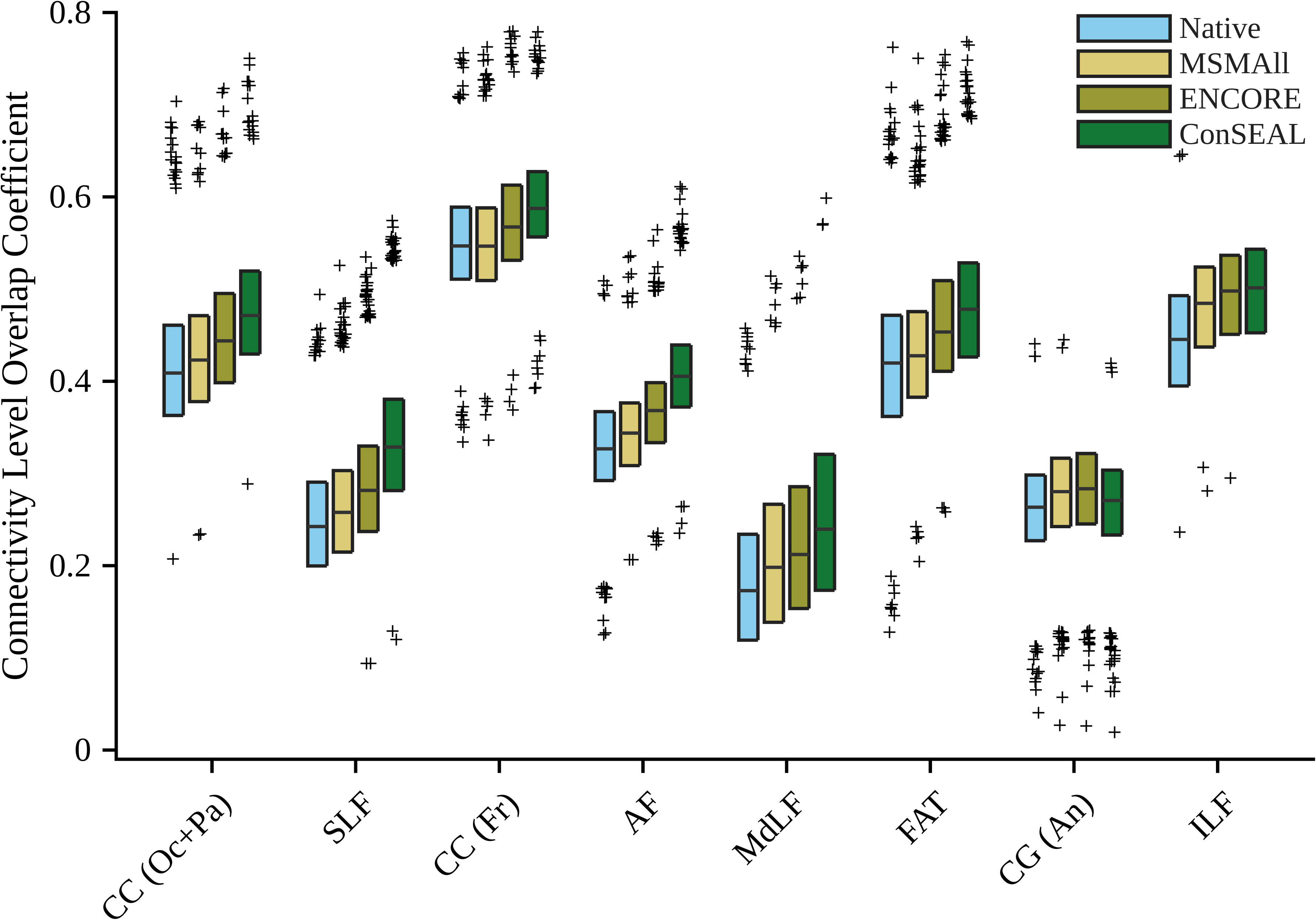}   \caption{Connectivity-level overlap coefficient for eight major fiber bundles.}
    \label{fig:connectivity_level_dice}
\end{figure}

We evaluate alignment accuracy using the connectivity-level overlap coefficient defined in \eqref{eqn:connectivity_dice}. Using 50 selected test subjects, we compare \textbf{ConSEAL} with raw unaligned streamline endpoint data (\textbf{Native}), \textbf{MSMAll} \citep{RobinsonMSM} and \textbf{ENCORE} \citep{cole2025alignment}. The connectivity-level overlap coefficient results for eight major fiber bundle tracts are shown in Figure~\ref{fig:connectivity_level_dice}. ConSEAL achieves higher overlap coefficients across most major fiber tracts, indicating improved connectivity-based alignment. Note that the overlap coefficient after alignment is also influenced by the number of streamlines within each tract of interest, as well as the distribution of their original endpoints. The MMD coefficient result is provided in Supplementary Section~S4. As the connectivity-level overlap coefficient evaluates connectivity-level alignment, the MMD coefficient provides an additional validation of the distribution-level discrepancy between streamline endpoint distributions. Here, we use only the streamline endpoints from the left hemisphere. The MMD coefficient is nonnegative, with larger values indicating greater discrepancy between distributions.

In the connectivity-level overlap coefficient evaluation reported in Figure~\ref{fig:connectivity_level_dice}, we consider the following major fiber bundles: Occipital and Parietal fibers of the Corpus Callosum (CC (Oc+Pa)), Superior Longitudinal Fasciculus (SLF), Frontal fibers of the Corpus Callosum (CC (Fr)), Arcuate Fasciculus (AF), Middle Longitudinal Fasciculus (MdLF), Frontal Aslant Tract (FAT), anterior portion of the Cingulum Bundle (CG (An)), and Inferior Longitudinal Fasciculus (ILF).
The commissural fibers, CC (Oc+Pa) and CC (Fr), connect the two hemispheres, whereas the association fibers, SLF, AF, MdLF, FAT, CG (An), and ILF, connect regions within the same hemisphere. Among the within-hemisphere bundles, AF, SLF, MdLF, and FAT are long-range language-related tracts; CG (An) is a limbic-related tract; and ILF is associated with the visual pathway. ENCORE achieves larger overlap coefficients than both the native and MSMAll methods.
Compared with ENCORE, ConSEAL achieves larger overlap coefficients for all fiber bundles except CG (An).  Moreover, from the MMD results in Supplementary Figure~S1, we see that ConSEAL achieves better results for all fiber bundles on average than ENCORE, native and MSMAll.  

\subsection{Population-Level Analysis}
\label{sec:population}
When the endpoints of fiber tracts from different subjects are aligned to a common template subject, we expect inter-subject variability to decrease. To evaluate this effect at the population level, we use the Destrieux atlas \citep{destrieux2010automatic} as the reference parcellation. The Destrieux atlas comprises 148 cortical parcels (74 per hemisphere). 
For each subject, we construct a parcel-to-parcel connectivity matrix by counting the number of streamlines connecting every pair of parcels.

\begin{figure}[ht]
    \centering
    \includegraphics[width=0.9\columnwidth]{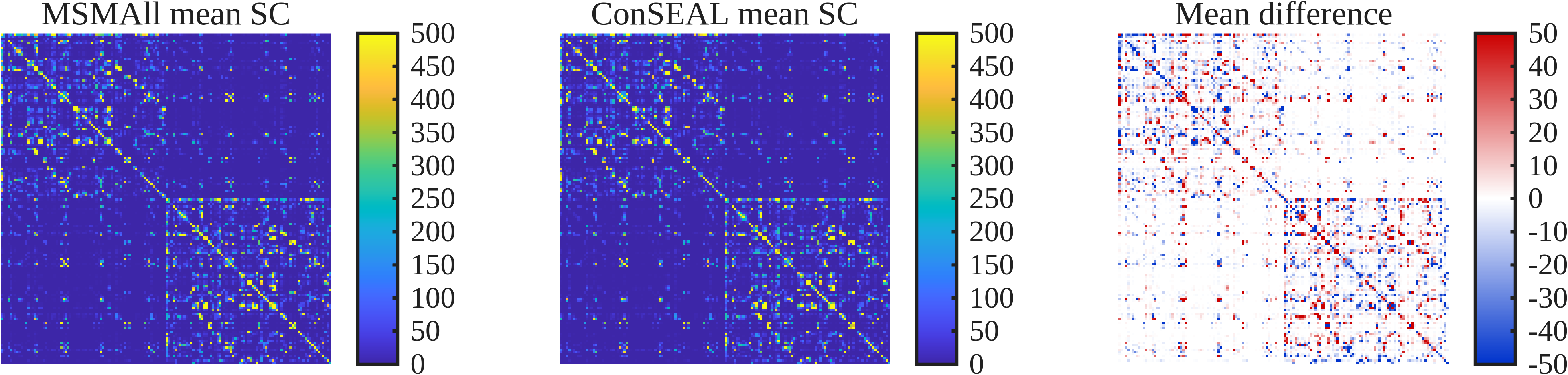}\\[0.5em]
    \includegraphics[width=0.9\columnwidth]{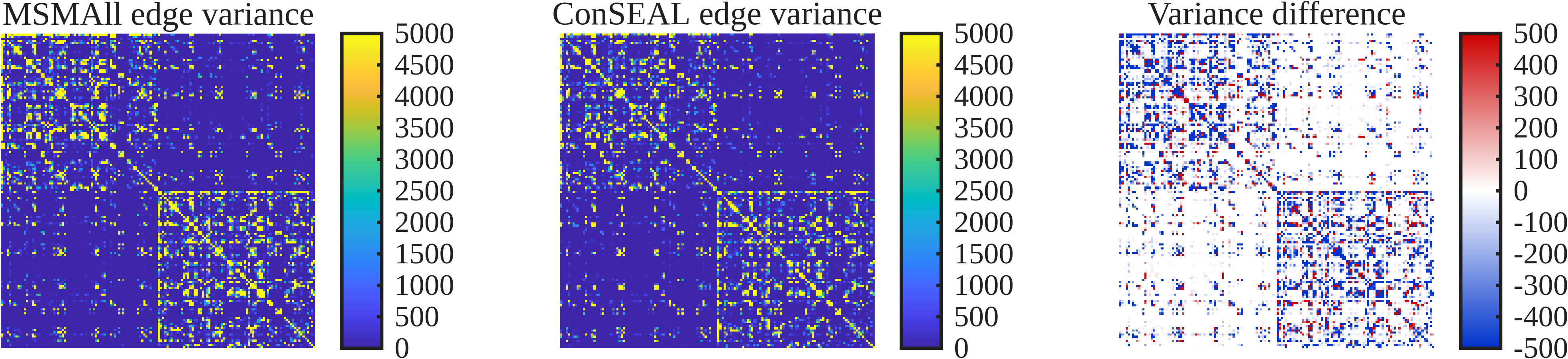}
    \caption{Atlas-level connectivity comparison before and after ConSEAL registration. The top row shows the mean number of fiber tracts connecting each pair of parcels, averaged across 80 subjects. The bottom row shows the corresponding inter-subject variance. In both rows, the third plot displays the difference between the ConSEAL-registered and MSMAll-registered data (ConSEAL $-$ MSMAll), illustrating the impact of registration on connectivity estimates and their variability.}
    \label{fig:atlas_level_variance}
\end{figure}

Figure~\ref{fig:atlas_level_variance} summarizes the atlas-level results based on 80 randomly selected subjects. The top row shows the mean streamline count for each parcel pair, averaged across subjects. The bottom row shows the corresponding inter-subject variance. In each row, the third panel shows the difference between ConSEAL and MSMAll registration, computed as ConSEAL $-$ MSMAll.

In the variance-difference matrix, 63.93\% of parcel pairs showed reduced variance after ConSEAL registration, while only 0.01\% showed exactly zero variance difference. Across parcel pairs, the median variance reduction was 27.21\% (IQR = 94.12\%). A paired sign-flip permutation test indicated that the variance differences were shifted below zero ($p < 1 \times 10^{-4}$, 10,000 permutations), and this result was also supported by a Wilcoxon signed-rank test ($p = 1.84 \times 10^{-216}$). These results indicate a broad reduction in inter-subject variance of fiber tract counts after ConSEAL registration. To assess whether this variance reduction could be explained by trivial attenuation of connectivity strength, we examined the mean-fiber-count differences. The total mean fiber count was preserved between the two matrices, indicating no global loss of connectivity strength. At the edge level, mean fiber counts increased for 43.12\% of parcel pairs and decreased for 56.87\%, with 0.01\% unchanged; the median percent change was -8.19\% (IQR = 59.56\%). The two-sided paired sign-flip permutation test on the mean difference was not significant ($p = 0.229$), consistent with preservation of the average mean connectivity across parcel pairs. Thus, the observed variance reduction is unlikely to be explained by global attenuation of mean connectivity, although ConSEAL redistributes mean fiber counts across edges.

\begin{figure}[ht]
    \centering
    \includegraphics[width=\columnwidth]{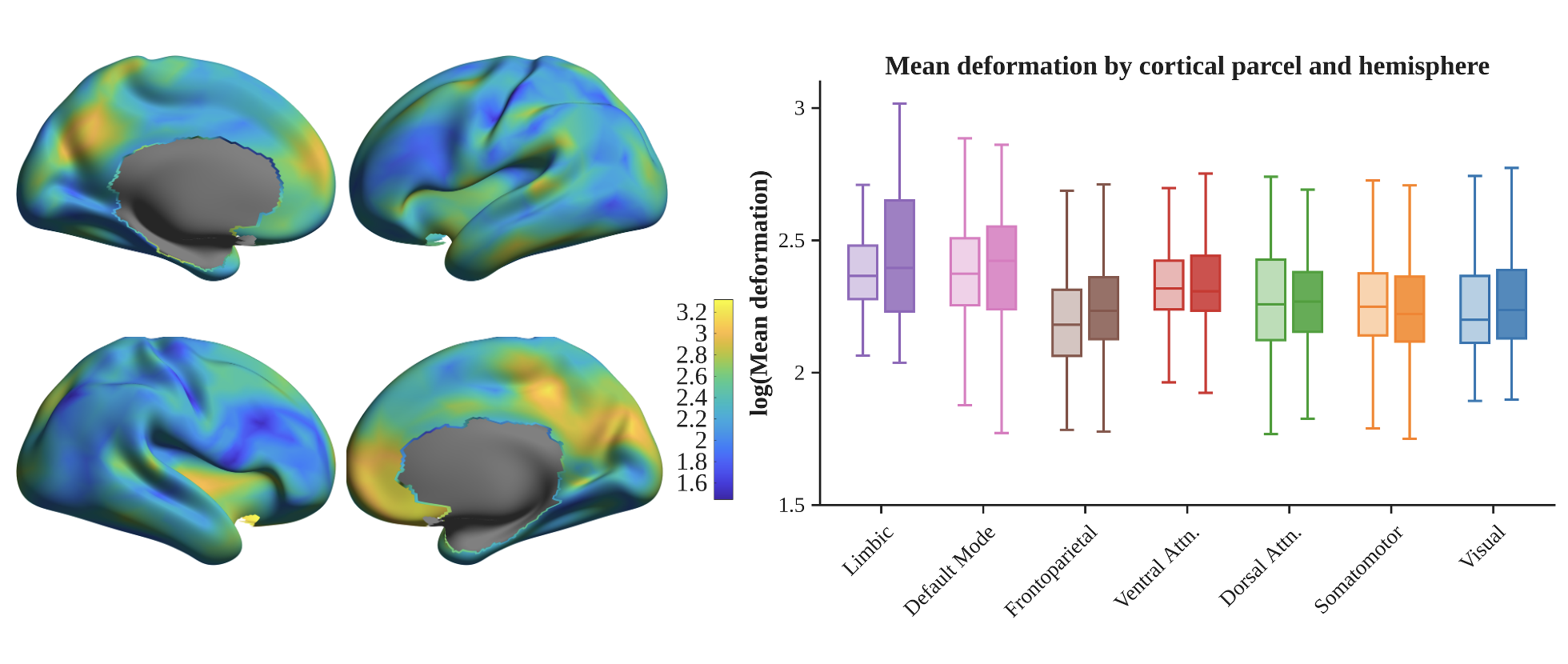}
    \caption{\textbf{Left:} Log-scale representation of the mean deformation field after centering the estimated warpings by their Karcher mean. \textbf{Right: } Boxplots of the log-scale mean deformation across parcels defined by the Yeo-7 network parcellation.}
    \label{fig:mean_deformation_log}
\end{figure}

We also visualize the mean deformation amplitude across subjects using the same 80 subjects included in the atlas-level connectivity comparison. Since all subjects are registered to a common template, we first compute the Karcher mean of the warpings estimated by ConSEAL. We then compose each individual warping with the inverse of this mean transformation. As a result, the transformed warpings have a Karcher mean equal to the identity map. In Figure~\ref{fig:mean_deformation_log}, the left panel shows the corresponding log-scale deformation amplitude field, and the right panel presents boxplots of the log-scale mean deformation stratified by the Yeo-7 network parcellation \citep{yeo2011organization}. Overall, the Limbic and Default Mode networks exhibit the largest deformations under ConSEAL, whereas unimodal sensorimotor and visual networks deform the least. This spatial pattern is anatomically expected. Inter-individual variability in both cortical folding and functional-connectional organization is greatest in heteromodal association cortex---particularly the default-mode and limbic systems---and lowest in primary sensorimotor and visual cortex \citep{mueller2013individual,gratton2018functional}. In precisely these high-variability association regions, folding geometry is most weakly coupled to areal and connectional organization, so folding- and MSMAll-based alignment leave the largest residual connectivity misregistration \citep{glasser2016multi}. ConSEAL must therefore apply its largest warps there to bring homologous connections into correspondence, while well-aligned sensory regions require little adjustment. The concentration of deformation in the association cortex thus indicates where connectivity-driven alignment contributes most beyond geometry-based registration, rather than reflecting spurious or unconstrained warping.

\begin{figure}[ht]
    \centering
    \includegraphics[width=\columnwidth]{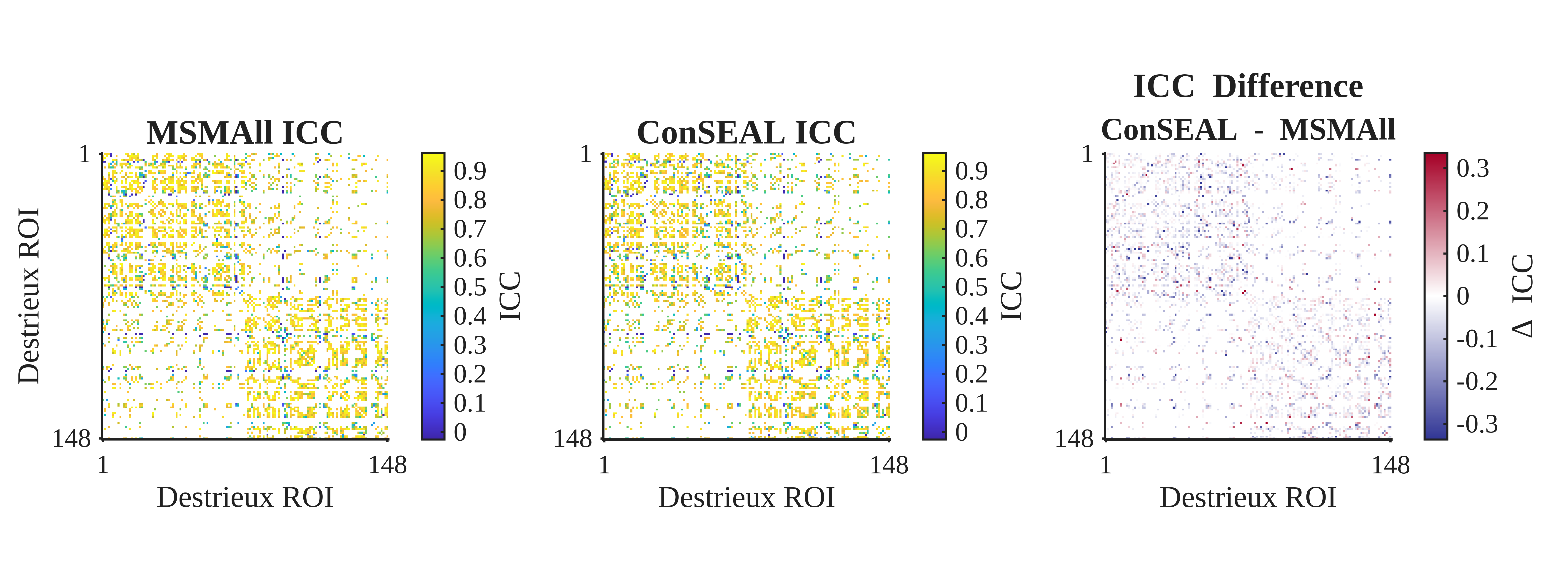}
    \caption{Edge-wise test--retest reliability across Destrieux cortical ROI pairs. The left and middle panels show intraclass correlation coefficient (ICC) matrices for MSMAll and ConSEAL alignment, respectively. The right panel shows the edge-wise ICC difference computed as ConSEAL minus MSMAll, where positive values indicate higher reliability after ConSEAL alignment and negative values indicate higher reliability with MSMAll. Rows and columns correspond to Destrieux ROI indices.}
    \label{fig:icc}
\end{figure}

We then evaluated test--retest reliability using the intraclass correlation coefficient (ICC). ICC quantifies the proportion of total variance attributable to between-subject differences rather than within-subject variability; higher ICC values therefore indicate greater reliability and individual identifiability. Thus, higher ICC values indicate that connectivity patterns are more consistent within subjects and more distinguishable across subjects. We used 37 test--retest subject pairs from the HCP dataset to evaluate reliability.
Using the Destrieux atlas as the reference parcellation, we count the number of fiber tracts connecting every pair of parcels. We exclude edges with low fiber counts under both MSMAll and ConSEAL alignment to reduce the influence of sparse, unstable connections. ICC is then computed over the remaining edges to assess whether ConSEAL alignment improves the reliability and individual identifiability of structural connectivity.
Figure~\ref{fig:icc} shows the edge-wise ICC matrices for MSMAll alignment, ConSEAL alignment, and their difference. ConSEAL alignment does not lead to a substantial increase in ICC, likely because ConSEAL reduces both within-subject and between-subject differences.

\begin{figure}[ht]
    \centering
    \includegraphics[width=0.9\columnwidth]{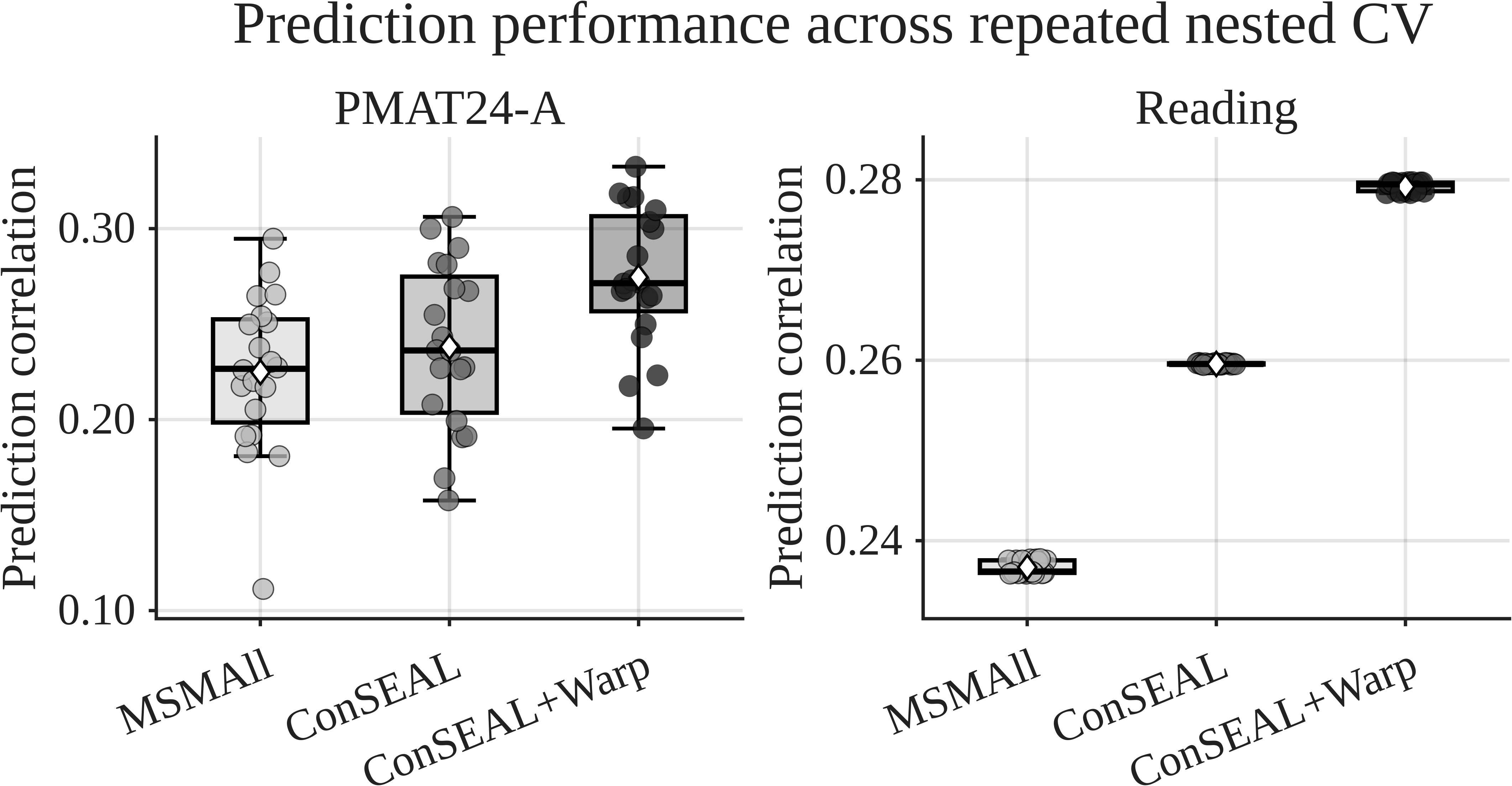}
    \caption{Prediction performance for PMAT24-A and Reading across 20 repeated nested cross-validation runs. Boxplots show prediction correlations for MSMAll-aligned tract-count features, ConSEAL-aligned tract-count features, and ConSEAL-aligned features augmented with warp information.}
    \label{fig:prediction}
\end{figure}

To evaluate the downstream utility of ConSEAL, we predicted two HCP cognitive traits: PMAT24-A and Reading. PMAT24-A is the Penn Matrix Reasoning Test, a nonverbal reasoning measure that requires participants to solve matrix-pattern problems based on geometric analogy and contrast. Reading is the NIH Toolbox Oral Reading Recognition score, a measure of reading decoding and crystallized ability based on accurate pronunciation of letters and words.
We compared three feature representations: MSMAll-aligned tract-count features, ConSEAL-aligned tract-count features, and ConSEAL-aligned features augmented with warp information. All features were derived from fiber tract counts using the Destrieux atlas. Prediction performance was assessed using the first 250 subjects with repeated nested cross-validation, robust preprocessing, and partial least squares (PLS) regression. As shown in Figure~\ref{fig:prediction}, ConSEAL improves prediction correlation for both traits relative to MSMAll, and adding warp information yields the highest performance.

\subsection{Computational Cost}
\label{sec:computational_complexity}

Because ConSEAL is robust to grid resolution (Section~\ref{robust_to_resolution}), accurate alignment is obtained on a coarse icosphere-4 grid, which keeps the cost moderate; the truncated, spatially cut-off spherical heat kernel further yields a sparse kernel matrix (Supplementary Section~S2). Table~\ref{tab:runtime_breakdown} reports the per-step runtime breakdown on CPU and GPU. The KDE evaluation dominates the runtime and scales quadratically with the number of grid vertices ($O(V^2)$), making it the primary bottleneck; GPU acceleration reduces all steps substantially. With $\sigma = 0.005$, $\delta = 0.1$, and $\epsilon = 10^{-6}$, ConSEAL converges in roughly 60 iterations on icosphere~4 and 450 on icosphere~5. Two strategies that further reduce the KDE cost---evaluating it once every $k$ iterations and a coarse-to-fine multiresolution schedule---are described in Supplementary Section~S7.

\begin{table}[ht]
\centering
\renewcommand{\arraystretch}{1.15}
\caption{Runtime breakdown (\%) for different icosphere resolutions and execution settings. Percentages are relative to the total runtime and may not sum to exactly 100\% due to rounding. Values in parentheses indicate the average runtime per iteration.}
\label{tab:runtime_breakdown}
\begin{tabular}{llccc}
\toprule
\textbf{Setting} & \textbf{Resolution} & \textbf{Gradient Estimation} & \textbf{Endpoint Update (Warp)} & \textbf{KDE Evaluation} \\
\midrule
CPU & icosphere 4 & 26.98\% (0.9014s) & 19.79\% (0.6612s) & 56.62\% (1.7790s) \\
CPU & icosphere 5 & 29.74\% (12.9725s) & 1.33\% (0.5817s) & 68.92\% (30.0631s) \\
GPU & icosphere 4 & 0.51\% (0.0093s) & 40.73\% (0.7402s) & 58.76\% (1.0677s) \\
GPU & icosphere 5 & 1.23\% (0.1365s) & 7.66\% (0.8501s) & 91.11\% (10.1121s) \\
\bottomrule
\end{tabular}
\end{table}

\section{Discussion}
\label{sec:discussion}

ConSEAL aligns cortical surfaces by directly warping streamline endpoints, and across major fiber bundles it achieves higher connectivity-level overlap than folding- and connectivity-based baselines. Its significance for network neuroscience, however, lies less in registration accuracy per se than in what that accuracy buys at the population level: by aligning the actual wiring, ConSEAL reduces the cross-subject variance of parcel-wise structural connectivity relative to MSMAll (Section~\ref{sec:population}), so that group-level connectomes become more consistent and, in principle, better powered to detect genuine individual and group differences.

A natural concern is that this variance reduction could be circular or trivial: ConSEAL optimizes an endpoint objective, and an endpoint-derived quantity (the SC matrix) then becomes less variable; moreover, variance can always be reduced by attenuating signal. Three observations argue against a trivial explanation. First, the reduction is not driven by global attenuation: the mean parcel-wise connectivity is preserved (the paired test on mean differences is non-significant), so ConSEAL redistributes rather than washes out connectivity. Second, the benefit transfers to a quantity ConSEAL never optimizes: relative to MSMAll, ConSEAL-aligned connectomes predict matrix reasoning (PMAT24-A) and oral reading more accurately, with a further gain when the estimated warps are added as features. Improved prediction is difficult to reconcile with the mere erasure of subject-specific signal. Third, the deformations are anatomically structured rather than diffuse: as shown above (Section~\ref{sec:population}), they concentrate in association cortex---the Limbic and Default Mode networks, where geometry-based alignment is weakest---rather than spreading uniformly, as a variance-minimizing artifact would.

Consistent with this interpretation, edge-wise test--retest reliability (ICC) is essentially unchanged between MSMAll and ConSEAL. We read this as expected rather than disappointing: because ICC is the ratio of between-subject to total variance, and ConSEAL reduces both within- and between-subject variance, the ratio is largely preserved. Individual identifiability is therefore not degraded---ConSEAL removes nuisance variance without erasing the subject-specific structure that supports reliability and prediction. Establishing a net gain in reliability, as opposed to its preservation, would likely require a dedicated retest cohort and is left to future work.

Methodologically, ConSEAL differs from prior connectivity-based registration in two ways. By using the continuous density only to generate a gradient and applying every update to the exact endpoint coordinates, it avoids the repeated grid interpolation and Jacobian estimation that introduce numerical diffusion, and its accuracy depends only weakly on grid resolution (Section~\ref{robust_to_resolution}). And because each incremental warp is the exponential map of a smooth tangent field on the sphere, the output is diffeomorphic by construction, without explicit smoothness or invertibility penalties. The price is computational: the spherical KDE dominates runtime, and although the every-$k$-iteration and multiresolution schemes (Supplementary Section~S7) reduce this cost, ConSEAL remains slower at inference than amortized deep-learning registration \citep{balakrishnan2019voxelmorph,zhao2021s3reg}. Because ConSEAL provides a geometrically principled, diffeomorphism-guaranteeing objective, it is a natural target for integration into a learned framework that retains these guarantees while improving speed.

ConSEAL also offers flexibility in what is aligned. Restricting the KDE, gradient, and deformation to the endpoints of a chosen set of tracts yields substantially higher overlap for those tracts; we nonetheless align all streamline endpoints, because endpoints outside the tracts of interest regularize the optimization and prevent large, unrealistic deformations in otherwise unconstrained regions. The diffusion bandwidth $\sigma$ is selected by maximizing bundle overlap (Section~\ref{sec:hyper_select}) and is corroborated by leave-one-out cross-validation, which returns the same value $\sigma = 0.005$ (Supplementary Section~S6).

Several limitations remain. Our evaluation uses a single cohort of healthy young adults (HCP) and depends on the quality of SET tractography and RecoBundlesX bundle segmentation, both of which carry known biases \citep{maier2016tractography}; the population-level analyses, although mutually consistent, rest on this single dataset. Future work includes validation on independent and clinical cohorts, joint structure--function alignment, a dedicated test--retest evaluation of reliability and identifiability, and integrating ConSEAL's diffeomorphic objective with deep-learning registration to combine guaranteed topology with fast inference.

\vspace{6pt} 





\authorcontributions{Y.X.: Conceptualization, Methodology, Validation, Formal analysis, Investigation, Data curation, Writing---original draft preparation, Visualization; M.C.: Conceptualization, Methodology, Software, Data curation; Z.Z.: Conceptualization, Methodology, Resources, Writing---review and editing, Supervision, Project administration. All authors have read and agreed to the published version of the manuscript.}

\funding{This research received no external funding.}

\institutionalreview{Not applicable.}

\informedconsent{Not applicable.}

\dataavailability{The data used in this study were obtained from the Human Connectome
Project (HCP) database and are publicly available at
\url{https://www.humanconnectome.org/study/hcp-young-adult}.
Access to the data requires registration and acceptance of the HCP
Open Access Data Use Terms.
The source code used in this study is available at:
\url{https://github.com/MartyCole/Encore/}.}

\conflictsofinterest{The authors declare no conflict of interest.} 

\reftitle{References}


\bibliographystyle{apalike}
\bibliography{egbib}

\isPreprints{}{
\end{adjustwidth}
} 
\end{document}


\maketitle

\noindent This document collects the full alignment algorithm, the gradient derivation, the
spherical kernel-density-estimation (KDE) details, the leave-one-out bandwidth-selection
procedure, the acceleration strategies, and two supporting figures referenced in the main
text. Equation, figure, and section numbers are prefixed with ``S''. References of the form
``Eq.~(1)'' or ``Section~3'' without the ``S'' prefix point to the main manuscript.

\section{Gradient Computation}
\label{appendix:gradient_computation}
The gradient of the warping energy $H$ defined in Eq.~(4) of the main text, denoted by
$\nabla H$, is derived in the Supplement of \cite{cole2025alignment}. Let $\{b_i\}_{i\ge1}$ be
an orthonormal basis of the tangent space $T_{\gamma_{id}}(\Gamma_{\Omega})$ at $\gamma_{id}$.
Then $\nabla H$ can be approximated by a truncated expansion using the first $M$ basis
elements:
\begin{equation}
\label{eq:grad_expansion}
\nabla H \approx \sum_{i=1}^{M} (\nabla_{b_i}H)\,b_i.
\end{equation}

Now consider $\Omega = \mathbb{S}_1^2 \cup \mathbb{S}_2^2$, where the two spheres represent the left and right hemispheres. The diffeomorphism $\gamma$ is defined componentwise as
\begin{align}
\label{eqn:componentwise_on_two_sphere}
\gamma(x) =
\begin{cases}
\gamma^1(x) & \text{if } x \in \mathbb{S}_1^2, \\
\gamma^2(x) & \text{if } x \in \mathbb{S}_2^2.
\end{cases}
\end{align}

The directional derivative of the cost function \(H\) with respect to \(\gamma^1\) in the direction \(b_i \in T_{\gamma^1_{id}}(\Gamma_{\mathbb{S}^2})\) is given by
\begin{align}
\label{eq:optomega}
\nabla_{b_i} H_{\gamma^1} &= \int_{x \in \mathbb{S}^2_1} \int_{y \in \mathbb{S}^2_2} \Bigl\{ q_1(x,y) - q_2(x,y) \Bigr\} \notag \\
&\quad \times \left\{ \begin{bmatrix}\frac{\partial q_2(x,y)}{\partial x} & \frac{\partial q_2(x,y)}{\partial y}\end{bmatrix} \begin{bmatrix} b_i(x) \\ b_i(y) \end{bmatrix} + \frac{q_2(x,y)}{2} \Bigl( \nabla \cdot b_i(x) + \nabla \cdot b_i(y) \Bigr) \right\} dx\,dy \notag \\
&\quad + \int_{x \in \mathbb{S}^2_1} \int_{v \in \mathbb{S}^2_2} \Bigl\{ q_1(x,v) - q_2(x,v) \Bigr\} \notag \\
&\quad \times \left\{ \frac{\partial q_2(x,v)}{\partial x}\, b_i(x) + q_2(x,v)\,\nabla \cdot b_i(x) \right\} dx\,dv.
\end{align}
The derivative with respect to $\gamma^2$ is obtained analogously by exchanging the roles of the two hemispheres.

\section{Kernel Density Estimation}
\label{appendix:kde}
The truncated spherical heat kernel used in the kernel density estimate of the main text
(Eq.~(1)) is defined as
\begin{align}
\label{eq:kernel}
K_{\sigma}(x, y)=\sum_{l=0}^{H}(2l+1) e^{-l(l+1)\sigma} P_l(x \cdot y),
\end{align}
where $x\cdot y = \cos\theta$ is the cosine of the geodesic angle between $x$ and $y$ on the sphere, and $P_l$ denotes the Legendre polynomial of degree $l$. The bandwidth parameter $\sigma > 0$ controls the amount of diffusion, with larger $\sigma$ leading to greater smoothing through stronger attenuation of higher-frequency components via the factor $e^{-l(l+1)\sigma}$. The exact heat kernel is given by an infinite series; in practice, we truncate the expansion at $l = H$ for computational efficiency. This truncation is justified because the higher-order terms decay exponentially fast with $l$.

In the implementation, the kernel is first evaluated using the truncated spherical harmonic expansion. Entries corresponding to pairs with cosine similarity below a numerically determined threshold are then set to zero, resulting in a sparse kernel matrix. Equivalently, pairs of points whose geodesic distance exceeds a certain threshold---beyond which the kernel value is negligible---are ignored. This sparsification significantly reduces both computational cost and memory usage in subsequent matrix operations.

Note that applying a hard cutoff to the cosine similarity can make the numerical kernel approximation discontinuous in practice. However, this discontinuity is negligible when the cutoff threshold is chosen such that the kernel value at the truncation boundary falls below a prescribed numerical tolerance.

One advantage of the spherical heat kernel over the geodesic Gaussian kernel \citep{seguin2022connectome} is its spectral representation. From \eqref{eq:kernel}, the bandwidth parameter $\sigma$ appears only in the scalar weights $e^{-l(l+1)\sigma}$, while the geometric terms $P_l(x\cdot y)$ are independent of $\sigma$.

As a result, when evaluating the KDE for multiple values of $\sigma$, one can precompute the Legendre components $P_l(x\cdot y)$ once, and then update the kernel efficiently by rescaling each harmonic term via $e^{-l(l+1)\sigma}$. This avoids recomputing pairwise geodesic distances or kernel evaluations for each $\sigma$, and reduces the computation to simple weighted sums (or matrix multiplications in the discrete setting), leading to substantially improved computational efficiency.

\section{The Complete ConSEAL Algorithm}
\label{appendix:algorithm}
Algorithm~\ref{alg:gamma} gives the complete numerical procedure for ConSEAL, summarized in
the Methods of the main text. At each iteration the connectivity density of the moving subject
is re-estimated from the current endpoint locations by the spherical heat-kernel KDE
(main-text Eq.~(1)), an incremental warp is obtained from the truncated gradient of the
warping energy (Section~\ref{appendix:gradient_computation}), and this warp is applied
directly to the discrete endpoint coordinates. Because each incremental warp is the
point-wise Riemannian exponential of a smooth tangent vector field on $\mathbb{S}^2$, it is a
diffeomorphism for sufficiently small step sizes; the final transformation, a finite
composition of such increments, is therefore also a diffeomorphism.

\begin{algorithm}[ht]
\caption{ConSEAL: Direct Point-Cloud Cortical Alignment on $\Omega = \mathbb{S}_1^2 \cup \mathbb{S}_2^2$}\label{alg:gamma}
\begin{algorithmic}[1]
\State \textbf{Input}: Target endpoints $\{x_i\}$ and Source endpoints $\{y_j^{(0)}\}$
\State Initialize $k = 0$, cumulative warps $\gamma^{1(0)} = \gamma^{2(0)} = \gamma_{\mathrm{id}}$, step size $\delta$, tolerance $\epsilon$
\State Compute target connectivity $f_1$ from $\{x_i\}$ via the spherical heat kernel (main-text Eq.~(1)); obtain $Q$-transform $q_1 = Q(f_1)$
\Loop
    \State Compute moving connectivity $f_2^{(k)}$ from current endpoints $\{y_j^{(k)}\}$ via the spherical heat kernel (main-text Eq.~(1))
    \State Obtain $Q$-transform $q_2^{(k)} = Q(f_2^{(k)})$
    \State Compute the gradients $\nabla H_{\gamma^1}$ and $\nabla H_{\gamma^2}$ using $q_1$ and $q_2^{(k)}$ according to \eqref{eq:optomega}
    \If{$\|\nabla H_{\gamma^1}\|_{\mathbb{L}^2} < \epsilon$ \textbf{and} $\|\nabla H_{\gamma^2}\|_{\mathbb{L}^2}  < \epsilon$}
        \State \textbf{break}
    \EndIf
    \State Compute small incremental warps: $\delta\gamma^{1} \gets \exp_{\mathbb{S}^2}({\delta \nabla H_{\gamma^1}})$, $\delta\gamma^{2} \gets \exp_{\mathbb{S}^2}({\delta \nabla H_{\gamma^2}})$
    \State Form joint incremental warp $\delta\gamma = (\delta\gamma^{1}, \delta\gamma^{2})$
    \State \textbf{Direct Endpoint Update:} $\{y_j^{(k+1)}\} \gets \{\delta\gamma(y_j^{(k)})\}$
    \State Update cumulative tracking (optional): $\gamma^{(k+1)} \gets \gamma^{(k)} \circ \delta\gamma$
    \State $k \gets k + 1$
\EndLoop
\State \textbf{Output}: Aligned Endpoints $\{y_j^{(k)}\}$, Final Diffeomorphism $\gamma^{(k)}$
\end{algorithmic}
\vspace{0.2em}
\noindent \textit{Note: $\exp$ denotes the point-wise Riemannian exponential map on the base manifold $\mathbb{S}^2$, which acts as a retraction for sufficiently small step sizes $\delta$.}
\end{algorithm}

\section{Maximum Mean Discrepancy (MMD) Coefficient}
\label{appendix:MMD}
\begin{figure}[ht]
    \centering
    \includegraphics[width=0.8\columnwidth]{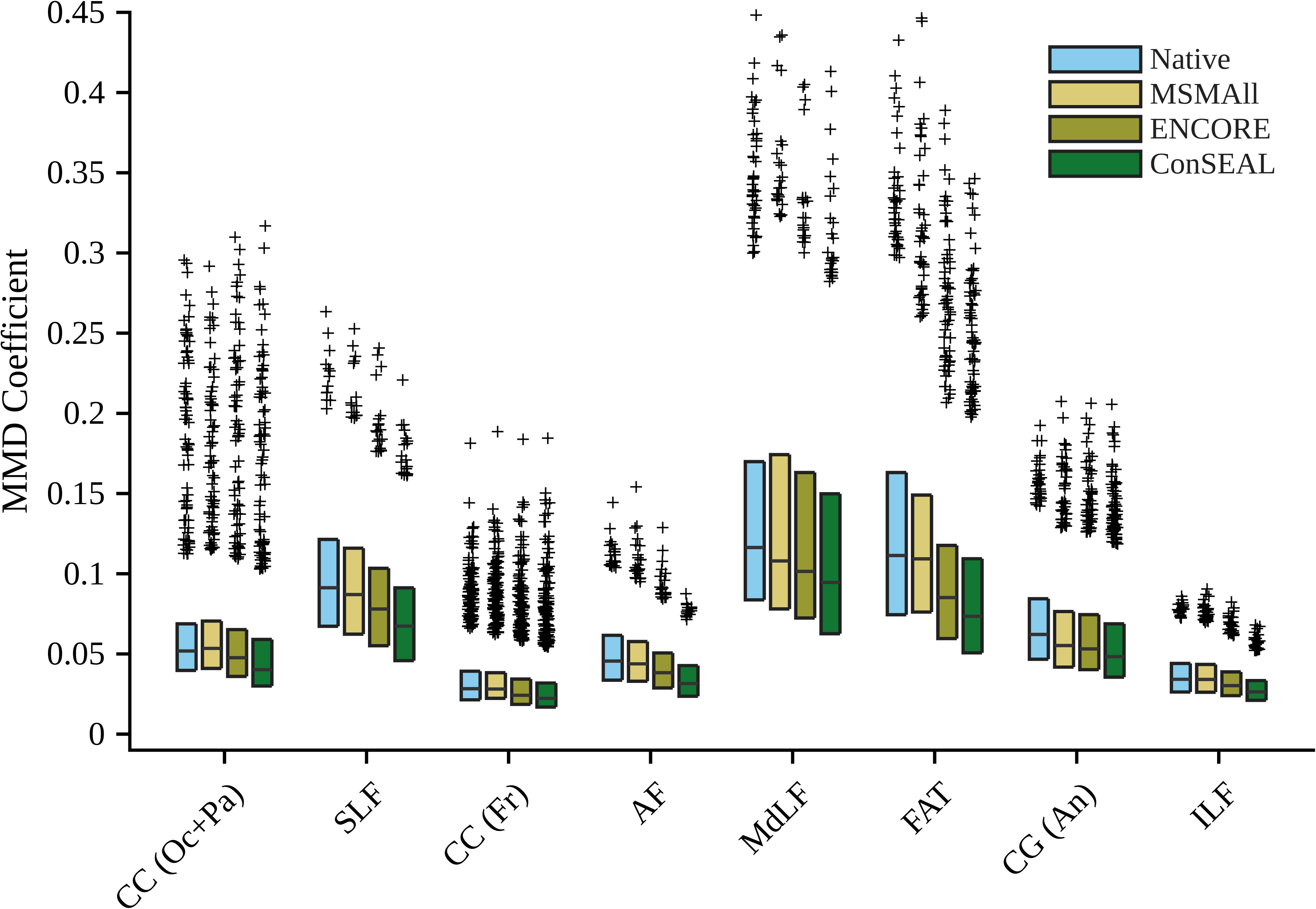}
    \caption{MMD coefficient for eight major fiber bundles.}
    \label{fig:MMD}
\end{figure}

The MMD coefficient measures the difference between two sample sets at the distribution level, capturing discrepancies in endpoint distributions rather than at the edge level. A lower MMD coefficient indicates that the two point clouds have more similar distributions. Figure~\ref{fig:MMD} reports the MMD coefficient for the eight major fiber bundles; on average ConSEAL attains the lowest MMD, indicating that it most effectively aligns the endpoints of major fiber bundles when they are viewed as point clouds. Here we use only the streamline endpoints from the left hemisphere.

\section{Ground-Truth Warping for the Robustness Experiment}
\label{appendix:additional_analysis}
\begin{figure}[h]
    \centering
    \includegraphics[width=0.45\linewidth]{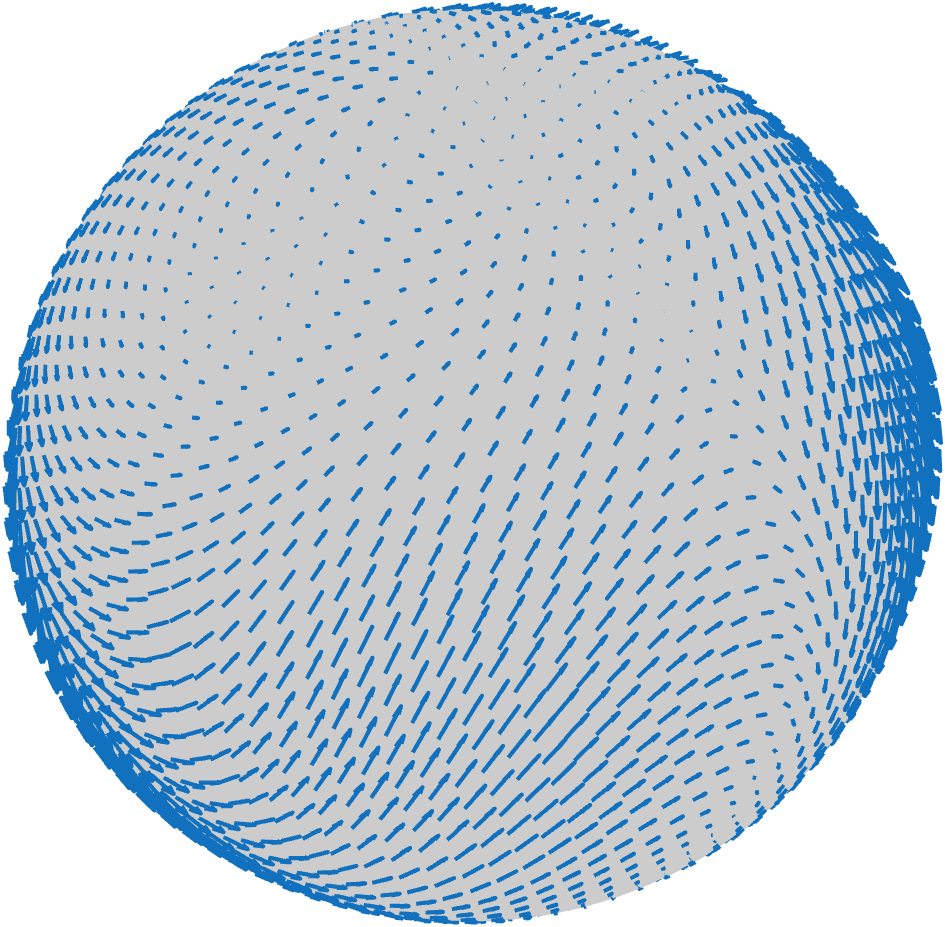}
    \caption{Ground-truth warping field used in the robustness experiment, visualized on an icosphere-4 mesh (2562 vertices).}
    \label{fig:ground_truth_warping}
\end{figure}

Figure~\ref{fig:ground_truth_warping} illustrates the ground-truth warping used in the grid-resolution robustness experiment of the main text.

\section{Bandwidth Selection via Leave-One-Out Cross-Validation}
\label{appendix:lcv}
In the main text we select the diffusion bandwidth parameter $\sigma$ using the
connectivity-level overlap coefficient of major fiber tracts. As an independent check, we also
performed a leave-one-out log-likelihood cross-validation (LCV) procedure based on the kernel
matrices. For each candidate $\sigma$, we construct a kernel-smoothed density estimate and
evaluate its predictive performance using leave-one-out likelihood. For each fiber endpoint
pair $(p_j^1,p_j^2)$, the contribution of the $j$-th fiber is removed from the density
estimate, and the resulting model is used to predict the connectivity at that fiber's
barycentric endpoints. The LCV score is defined as
\begin{align}
\operatorname{LCV}(\sigma)=\frac{1}{N} \sum_{j=1}^N \log \hat{f}_\sigma^{(-j)}\left(p_j^1, p_j^2\right),
\end{align}
where $\hat{f}_\sigma^{(-j)}$ denotes the leave-one-out density estimate obtained with the $j$-th fiber excluded. The optimal bandwidth is selected as the value of $\sigma$ that maximizes this score. This cross-validation procedure is computationally efficient within the KDE framework adopted in this paper (Section~\ref{appendix:kde}). The optimal bandwidth selected by this procedure was also $\sigma = 0.005$, in agreement with the overlap-coefficient criterion used in the main text.

\section{Acceleration Strategies}
\label{appendix:acceleration}
The runtime breakdown in the main text shows that the spherical KDE evaluation is the primary
computational bottleneck of ConSEAL. We describe two strategies that reduce this cost without
altering the qualitative update dynamics of the algorithm.

\paragraph{Every-$k$-iteration KDE.} The KDE can be evaluated once every $k$ iterations rather
than at every iteration. Specifically, the endpoints are updated (Endpoint Update) and the
continuous density proxy is recomputed (KDE Evaluation) only every $k$ steps. During the
intermediate iterations, ENCORE is used to directly warp the current density proxy without
recomputing the KDE. This strategy reduces the overall computational cost while preserving the
update dynamics of the algorithm.

\paragraph{Multiresolution schedule.} ConSEAL also supports a coarse-to-fine multiresolution
strategy. We first run Algorithm~\ref{alg:gamma} on a coarser grid (e.g., icosphere 3 with 642
vertices) and apply the resulting warping to the streamline endpoints. We then rerun
Algorithm~\ref{alg:gamma} on these updated endpoints using a finer grid (e.g., icosphere 4 with
2562 vertices), and compose the resulting warping with the previous one to obtain the final
transformation. In this multiscale framework, the bandwidth parameter $\sigma$ can also vary
across resolutions; for example, a larger $\sigma$ can be used on the coarse grid and gradually
reduced on finer grids. This approach accelerates convergence when the goal is to obtain a
warping on a very fine grid, since computations on the coarser grid are significantly less
expensive: large deformations are captured on the coarser grid and then refined on the finer
grid. The performance of this strategy depends on the choice of grid hierarchy, refinement
levels, and stopping criteria; poor choices may lead to suboptimal alignment.

\bibliographystyle{apalike}
\bibliography{egbib}